\documentclass{article} 
\usepackage{arxiv_conference,times}

\usepackage{amsmath,amsfonts,bm}

\def\eqref#1{equation~\ref{#1}}

\def\1{\bm{1}}

\DeclareMathAlphabet{\mathsfit}{\encodingdefault}{\sfdefault}{m}{sl}
\SetMathAlphabet{\mathsfit}{bold}{\encodingdefault}{\sfdefault}{bx}{n}

\usepackage{amsmath}
\usepackage{amssymb}
\usepackage{graphicx}
\usepackage{float}
\usepackage{booktabs}
\usepackage{arydshln}
\usepackage{array}
\usepackage{tabularx}
\usepackage{multirow}
\usepackage{xcolor}
\usepackage{pifont}
\usepackage{hyperref}
\usepackage{url}
\usepackage{wrapfig}
\usepackage[align=center,shadow=true,shadowsize=4.5pt,nobreak=true,framemethod=tikz,skipabove=9.5pt,skipbelow=9pt,innertopmargin=6pt,innerbottommargin=6pt,innerleftmargin=5pt,innerrightmargin=5pt,leftmargin=1.5pt,rightmargin=1.5pt]{mdframed}
\usetikzlibrary{shadows}

\graphicspath{{./}}
\newcolumntype{M}[1]{>{\centering\arraybackslash}m{#1}}
\newcommand{\best}[1]{\textbf{#1}}
\newcommand{\ours}{JustQuant}
\newcommand{\method}{Theseus QAD}
\newcommand{\oursbf}{\textbf{JustQuant}}
\newcommand{\methodbf}{\textbf{Theseus QAD}}

\title{JustQuant: You Don't Need Smoothing, SVD, or Rotation for 4-Bit Activation Quantization}
\vspace{-4mm}

\author{
  \textbf{%
  Kaicheng Yang$^{1,*}$,
  Kaisen Yang$^{2,*}$,
  Chunyu Liu$^{2,*}$,
  Xianglong Yan$^{1}$,
  Haotong Qin$^{3}$}\\
  \textbf{%
  Junyi Wu$^{1}$,
  Tianao Zhang$^{1}$,
  Xun Zhang$^{1}$,
  Shaoqiu Zhang$^{1}$,
  Youbang Sun$^{2,\dagger}$,
  Yulun Zhang$^{1,\dagger}$}\\
  $^1$Shanghai Jiao Tong University,$^2$Tsinghua University,$^3$The Hong Kong Polytechnic University\\
  $^*$Equal contribution. $^\dagger$Corresponding authors.\\
  \texttt{yangkaicheng@sjtu.edu.cn, \{yks23,liucy24\}@mails.tsinghua.edu.cn}\\
  \texttt{ybsun@mail.tsinghua.edu.cn, yulun100@gmail.com}
}

\begin{document}

\maketitle

\begin{figure}[H]
  \vspace{-12mm}
  \centering
  \setlength{\tabcolsep}{2pt}
  \renewcommand{\arraystretch}{1.0}

  \begin{tabular}{cccc}
    \begin{minipage}[t]{0.24\textwidth}
      \centering
      {\scriptsize FLUX.1-schnell (4 steps)\\
      \textcolor{red!70!black}{BF16}\\
      FID$\downarrow$ 19.15 / IR$\uparrow$ 0.959\\
      }

      \includegraphics[width=\linewidth]{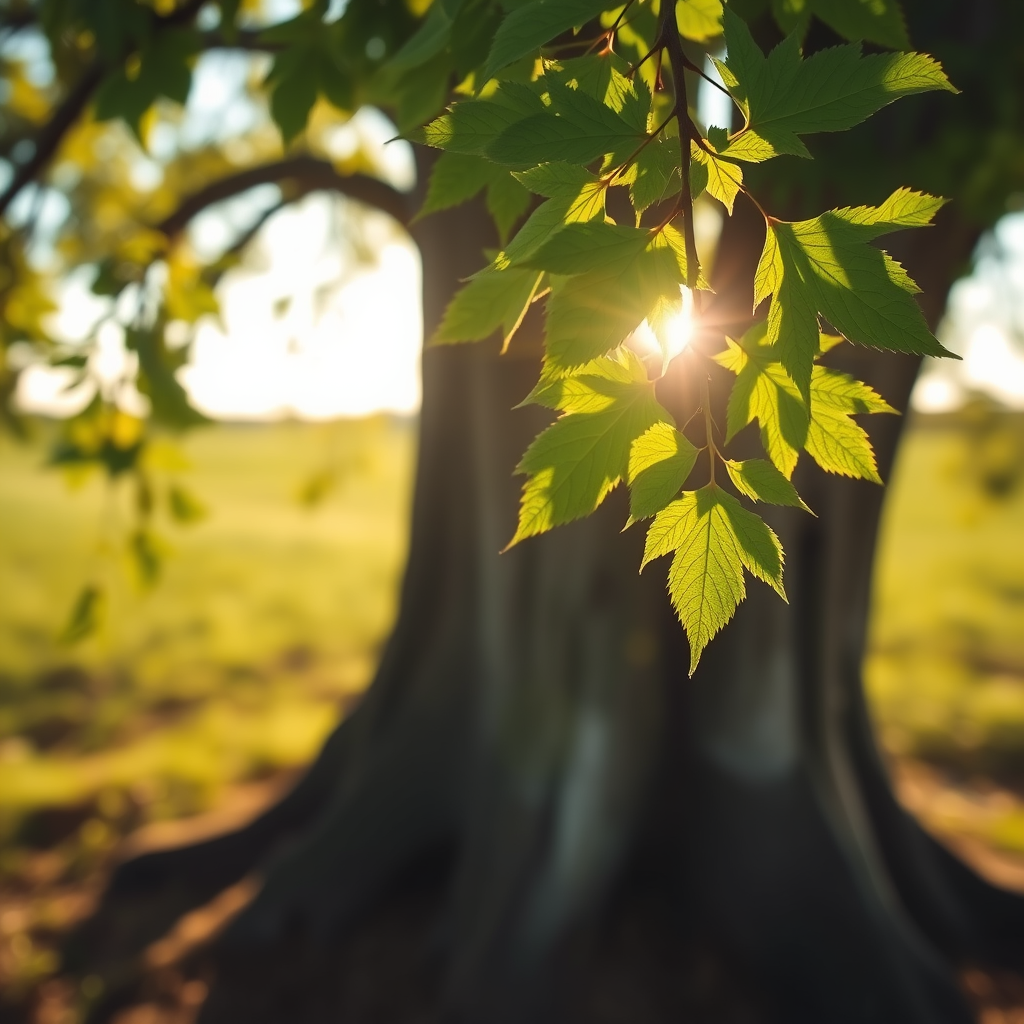}
    \end{minipage}
    &
    \begin{minipage}[t]{0.24\textwidth}
      \centering
      {\scriptsize ConvRot\\
      \textcolor{red!70!black}{W4A4+Rotation}\\
      FID$\downarrow$ 17.86 / IR$\uparrow$ 0.947\\
     }

      \includegraphics[width=\linewidth]{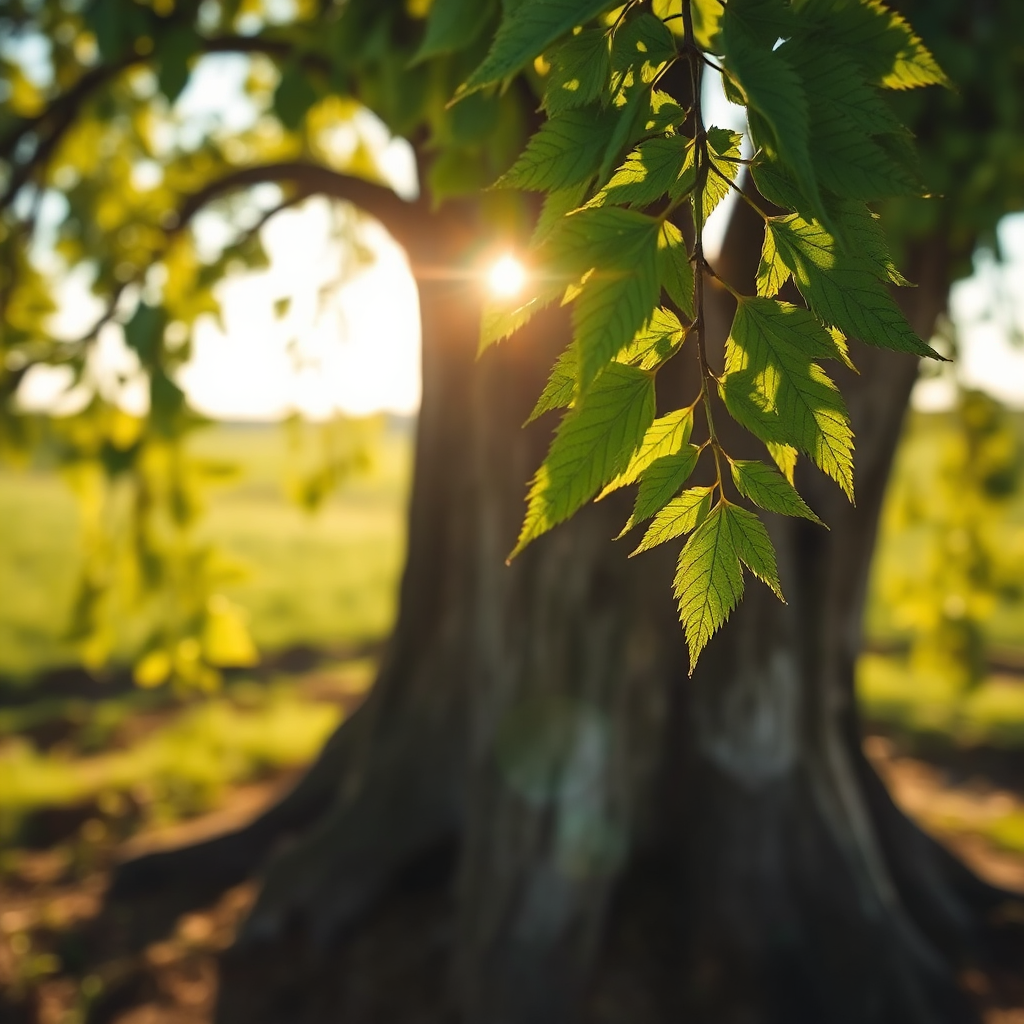}
    \end{minipage}
    &
    \begin{minipage}[t]{0.24\textwidth}
      \centering
      {\scriptsize SVDQuant\\
      \textcolor{red!70!black}{W4A4+Smooth+SVD}\\
      FID$\downarrow$ 18.51 / IR$\uparrow$ 0.968\\
      }

      \includegraphics[width=\linewidth]{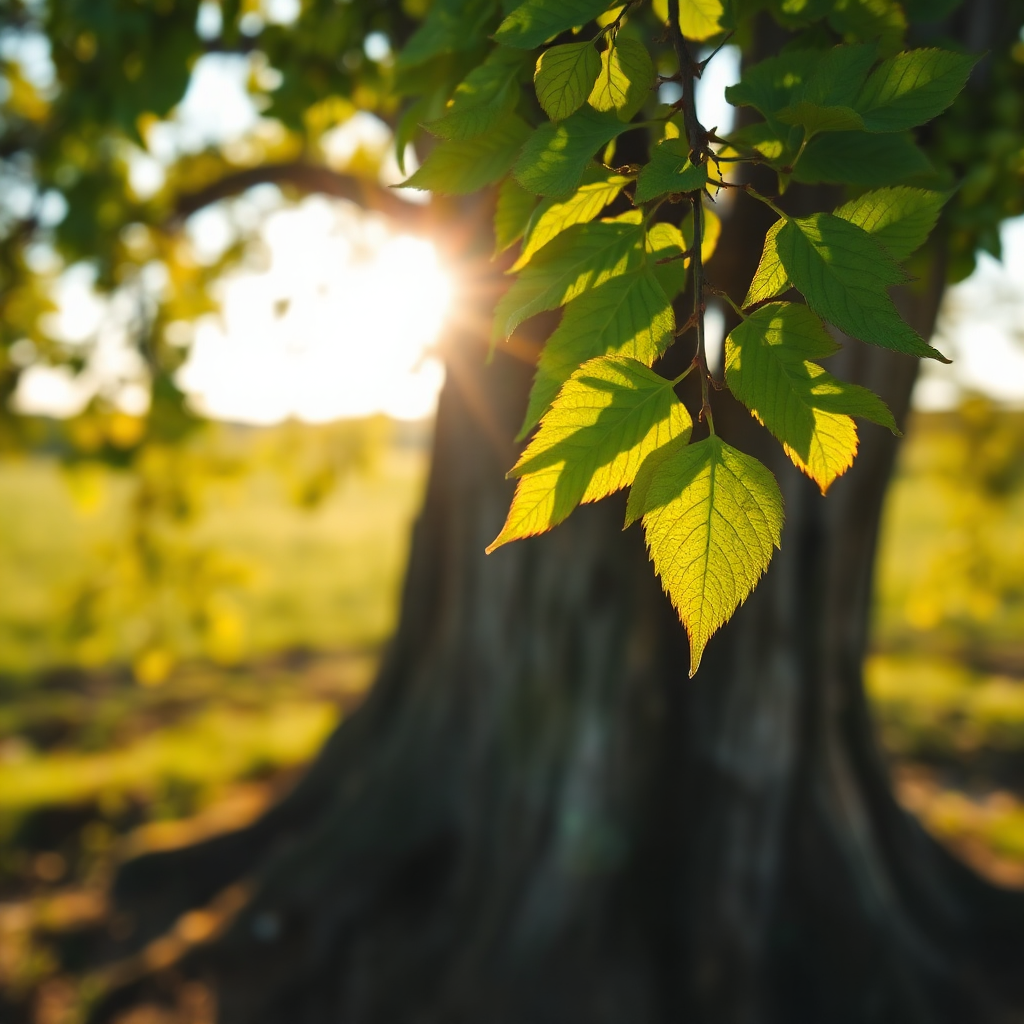}
    \end{minipage}
    &
    \begin{minipage}[t]{0.24\textwidth}
      \centering
      {\scriptsize \ours{}\\
      \textcolor{green!45!black}{Naive W4A4}\\
      FID$\downarrow$ 17.62 / IR$\uparrow$ 0.993\\
      }

      \includegraphics[width=\linewidth]{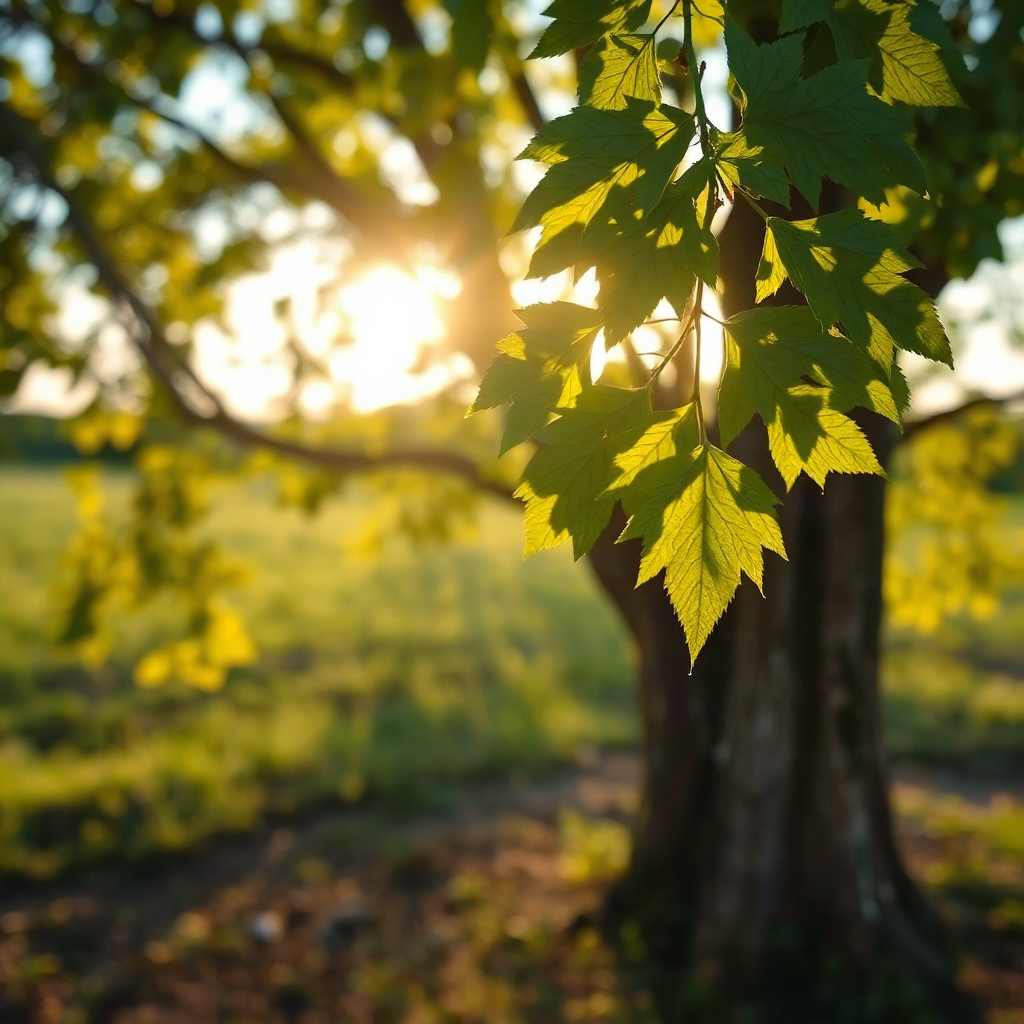}
    \end{minipage}
  \end{tabular}

  \vspace{0.5mm}

  \begin{tabular}{cccc}
    \begin{minipage}[t]{0.24\textwidth}
      \centering
      {\scriptsize FLUX.1-dev (50 steps)\\
      \textcolor{red!70!black}{BF16}\\
      FID$\downarrow$ 20.26 / IR$\uparrow$ 0.958\\
      }

      \includegraphics[width=\linewidth]{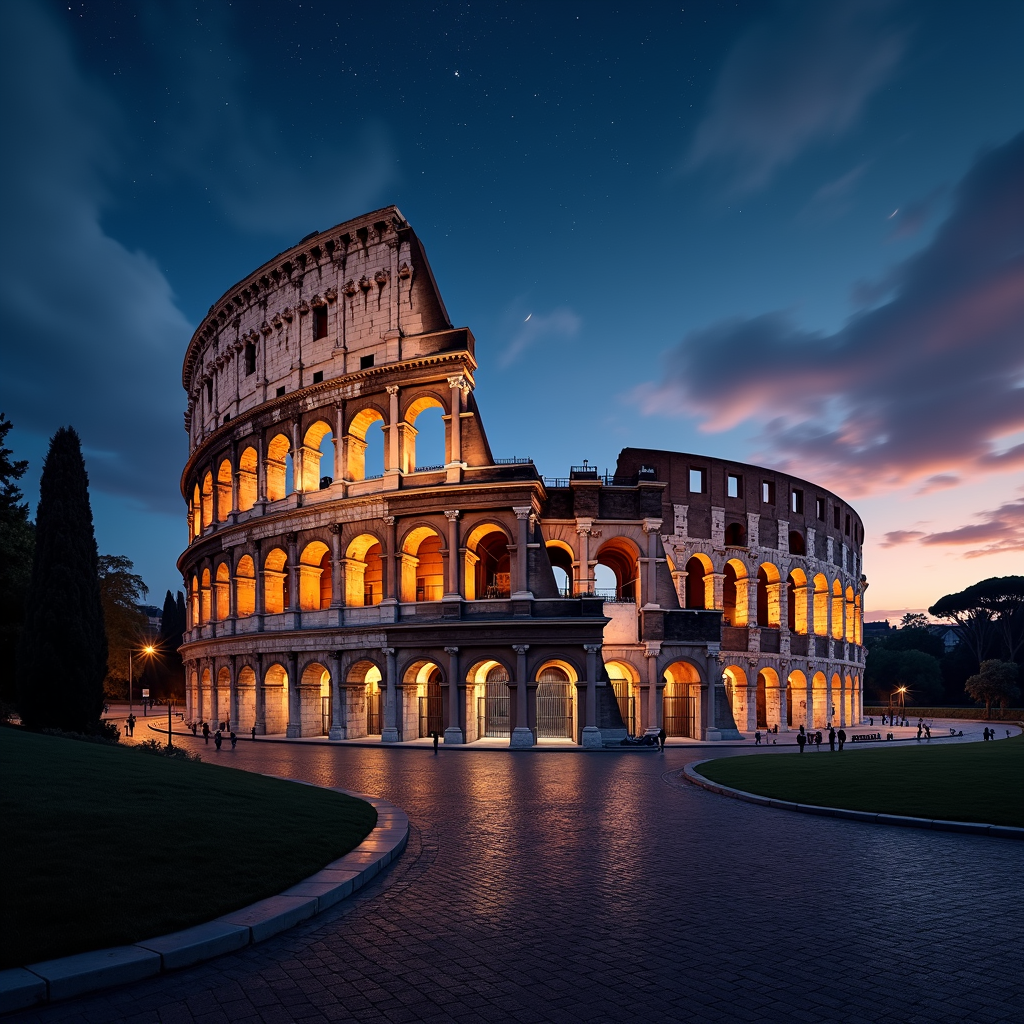}
    \end{minipage}
    &
    \begin{minipage}[t]{0.24\textwidth}
      \centering
      {\scriptsize PTQ\\
      \textcolor{green!45!black}{Naive W4A4}\\
      FID$\downarrow$ 24.29 / IR$\uparrow$ 0.754\\
      }

      \includegraphics[width=\linewidth]{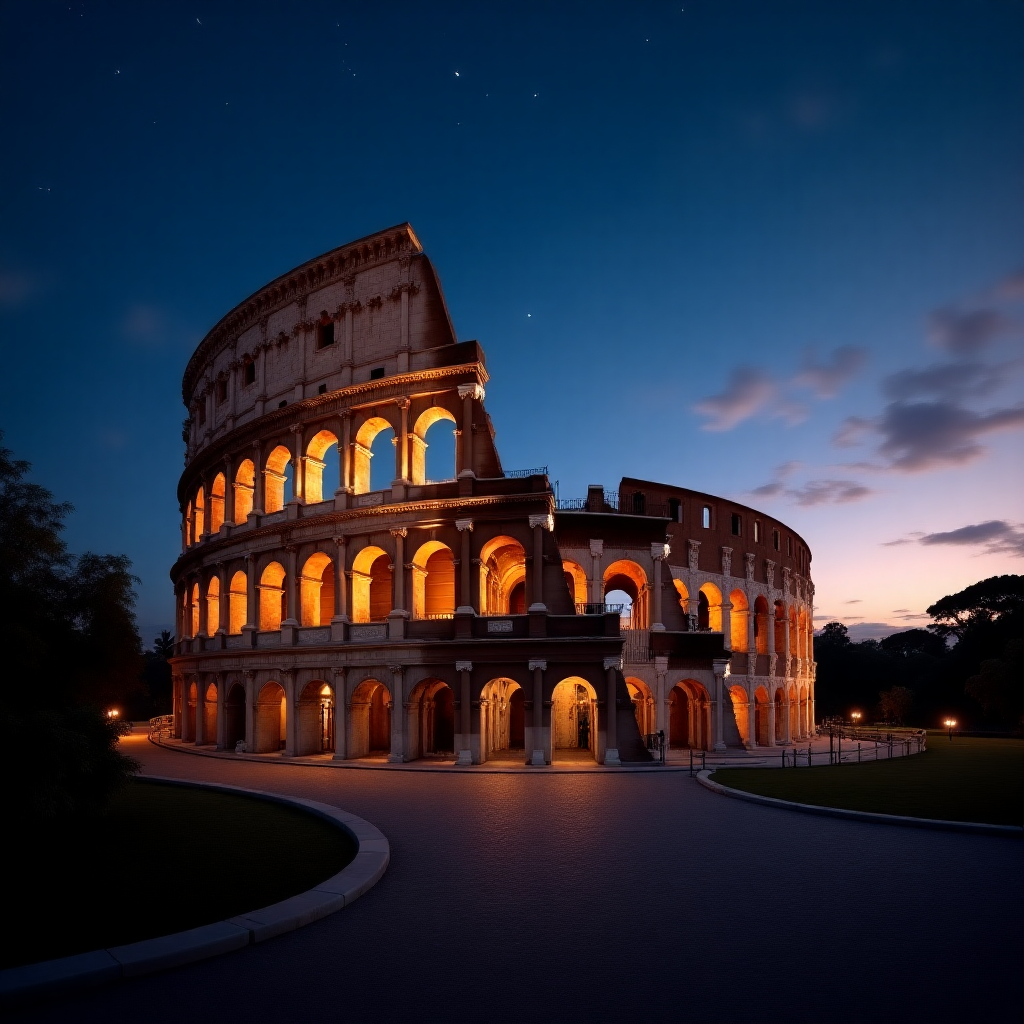}
    \end{minipage}
    &
    \begin{minipage}[t]{0.24\textwidth}
      \centering
      {\scriptsize QAT\\
      \textcolor{green!45!black}{Naive W4A4}\\
      FID$\downarrow$ 22.54 / IR$\uparrow$ 0.520\\
      }

      \includegraphics[width=\linewidth]{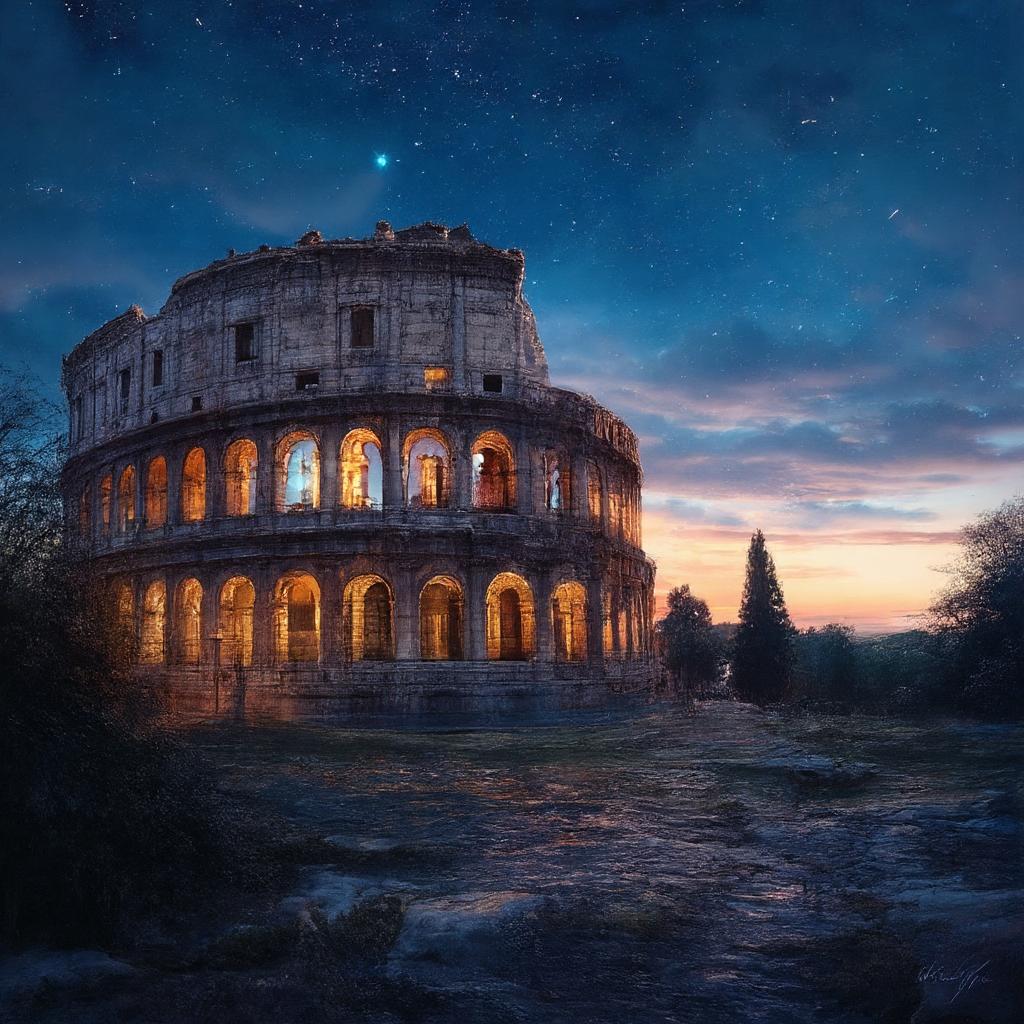}
    \end{minipage}
    &
    \begin{minipage}[t]{0.24\textwidth}
      \centering
      {\scriptsize \ours{}\\
      \textcolor{green!45!black}{Naive W4A4}\\
      FID$\downarrow$ 20.21 / IR$\uparrow$ 0.956\\
      }

      \includegraphics[width=\linewidth]{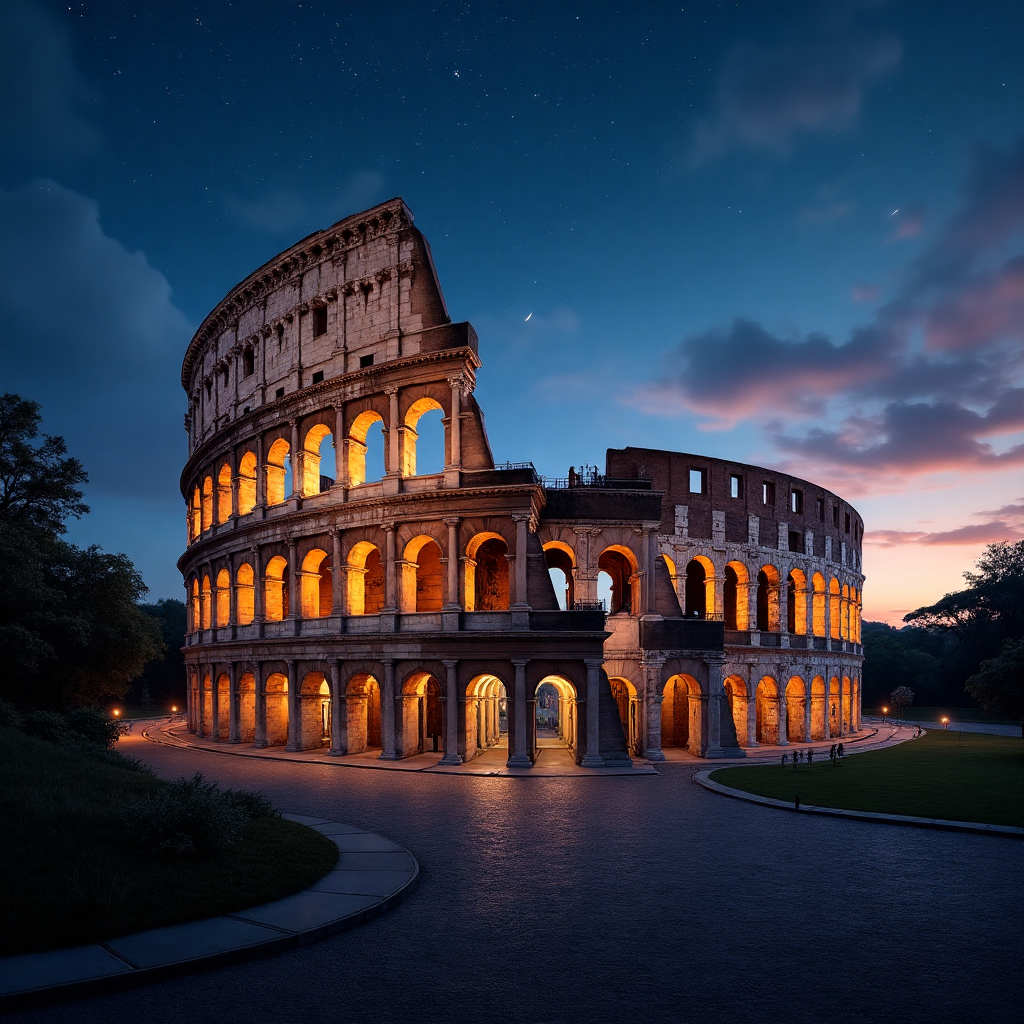}
    \end{minipage}
  \end{tabular}

 \vspace{-2mm}
  \caption{\textbf{FLUX-family 12B experiments.} \ours{} pushes naive W4A4 quality close to BF16 with only 18 hours on 4 H200 GPUs, while preserving a clean inference path.}
  \label{fig:justquant_overview}

  \vspace{-5mm}
\end{figure}

\begin{abstract}
\vspace{-3mm}
Recent generative models have become increasingly powerful, but their inference
cost continues to grow. Model quantization is a promising path for compressing
these models and accelerating inference. However, when quantization is pushed to
4 bits, activation quantization becomes substantially more difficult than weight
quantization. Recent PTQ (post-training quantization) and QAT
(quantization-aware training) methods have made progress on 4-bit activation
quantization by introducing smoothing, SVD branches, rotations, mixed precision,
or advanced formats such as NVFP. Yet these operators and data types impose
demanding requirements on inference engines and hardware, limiting the broad
adoption of low-precision models. \textbf{Can we quantize using only plain
low-bit operators?} To answer this question, we propose \oursbf{}, a simple yet effective
framework that moves the complexity required for low-bit quantization from
deployment-side operators into an acceptable training process. First, we revisit
model quantization from the perspective of knowledge distillation, and show that
a key reason existing PTQ and QAT methods fail is that they typically exploit
supervision at only a single level. We then introduce \methodbf{}
(quantization-aware distillation), a quantization-aware distillation method
that progressively applies multi-level supervision, analogous to the gradual
replacement process in the Ship of Theseus. Extensive experiments on DiT and
dLLM show two regimes. For smaller models, \method{} can serve as a light
warm-up stage that substantially improves QAT with plain operators, while naive
QAD can collapse in the same setting. For larger models, \method{} provides a
stronger distillation training path than ordinary QAD. Across these regimes,
\method{} improves plain quantization while avoiding the complex 
operators required by many PTQ methods.

\end{abstract}

\begin{figure*}[t]
\vspace{-4mm}
\centering
\includegraphics[width=\textwidth]{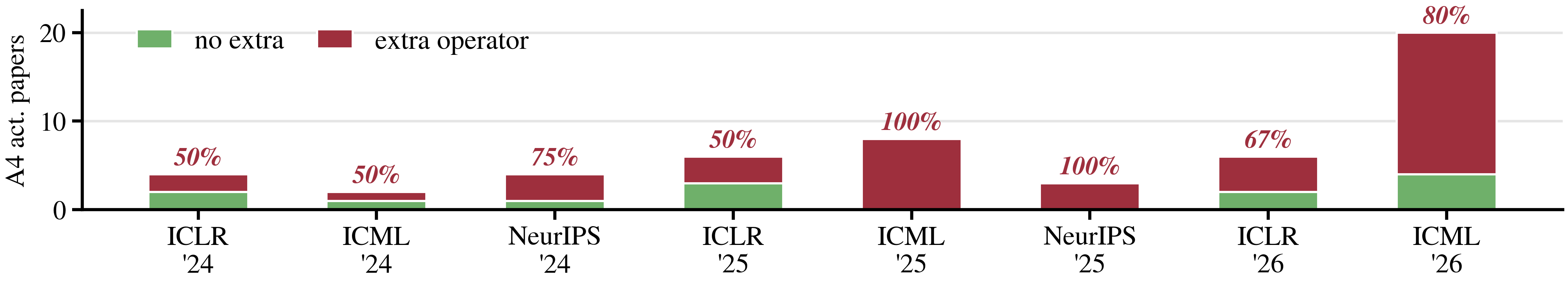}
\vspace{-8mm}
\caption{\textbf{The false prosperity of low-bit activation quantization.} Across eight recent conferences, $\leq$4-bit activation papers increasingly rely on extra operators.  This deployment-side complexity can obscure the practical speedups promised by low-bit activations.}
\label{fig:teaser}
\vspace{-5mm}
\end{figure*}
\section{Introduction}
\label{sec:introduction}
\vspace{-5mm}
\begin{wraptable}{r}{0.45\textwidth}
    \vspace{-5mm}
    \centering
    \setlength{\tabcolsep}{2.4pt}
    \renewcommand{\arraystretch}{1.10}
    \newcommand{\good}{\textcolor{green!45!black}{\ding{51}}}
    \newcommand{\fair}{\textcolor{yellow!70!black}{\textbf{--}}}
    \newcommand{\bad}{\textcolor{red!70!black}{\ding{55}}}
    \resizebox{\linewidth}{!}{%
    \begin{tabular}{lcccc}
        \toprule
        & \multicolumn{2}{c}{Small model} & \multicolumn{2}{c}{Large model} \\
        \cmidrule(lr){2-3}\cmidrule(lr){4-5}
        Method & Quality & Deploy. & Quality & Deploy. \\
        \midrule
        PTQ & \good & \bad & \good & \bad \\
        QAT & \good & \bad & \bad & \good \\
        QAD & \bad & \bad & \good & \good \\
        JustQuant & \good & \good & \good & \good \\
        \bottomrule
    \end{tabular}
    }
    \vspace{-2mm}
    \caption{\textbf{Different regimes favor different quantization recipes.}}
    \label{tab:regime_comparison}
    \vspace{-5.5mm}
\end{wraptable}
Recent 4-bit weight quantization methods can already compress generative models
with near-lossless quality and accelerate memory-bound inference scenarios by
reducing weight bandwidth~\citep{frantar2023gptq,lin2023awq,zhang2024lqer,li2024svdquant,fourtune2026}.
However, weight-only quantization does not
accelerate compute-bound transformer operations whose cost is dominated by
activation-weight matrix multiplication.  Without activation quantization, these
operations cannot be executed as fully low-bit GEMMs.  In diffusion transformers
and large text-to-image backbones, this cost is repeated across many denoising
steps~\citep{peebles2023dit,blackforestlabs2024flux}.  Pushing activations to 4
bits is therefore necessary for practical compute-bound acceleration, but it is
also substantially more difficult than weight quantization.  This bottleneck is repeatedly observed in recent work on DiT, video DiT, LLM, and FP4/MXFP4 quantization~\citep{feng2025qvdit,zhang2026qdit4sr,panferov2025quest,zhao2026twla,shao2025blockrotation}.  This is a fragile
regime for generative models: activation errors are injected into every
denoising or decoding step, shift internal feature distributions, reduce sample
diversity, and can destabilize training.

Existing 4-bit activation quantization methods follow two routes, both limited
for plain deployment.  PTQ methods often rely on smoothing, rotations, Hadamard
transforms, channel permutations, SVD branches, mixed precision, extra
normalization, or new formats such as NVFP4, imposing requirements on kernels
and hardware~\citep{xiao2023smoothquant,wu2024ptq4dit,chen2024qdit,zhao2024viditq,li2024svdquant,zhang2026qdit4sr,feng2025qvdit,ashkboos2024quarot,liu2024spinquant,saxena2025resq,huang2025slimllm,zhao2026twla,shao2025blockrotation,xin2026qadnvfp4}.
QAT appears more compatible with plain operators, but strong
results often still use rotation, SVD, NVFP4, or related inference-side
techniques~\citep{yang2025robuq,wang2025bitnetv2,panferov2025quest,fourtune2026}; without them,
plain W4A4 or W1.58A4 models often fail to reach usable quality.
Figure~\ref{fig:teaser} summarizes the growing reliance on extra operators
across recent conferences, highlighting the gap between low-bit formats and
practical deployment.  This raises
a question:

\vspace{-1.5mm}
\begin{mdframed}[
    leftmargin=0.2em,
    rightmargin=0.2em,
    linecolor=blue!65!black,
    backgroundcolor=blue!10!white,
    linewidth=1.2pt]
    \fontsize{10pt}{\baselineskip}\itshape
    \centering
    Can we achieve 4-bit activation quantization using only naive
    low-bit operators?
    \vspace{-.5mm}
\end{mdframed}
\vspace{-2.7mm}

We revisit model quantization from the perspective of \textbf{Net2Net knowledge
distillation}~\citep{du2024bitdistiller,feng2025qvdit,xin2026qadnvfp4}.  Because
the full-precision teacher and low-bit student share the same architecture,
quantization naturally exposes supervision signals at different granularities.
PTQ uses dense local reconstruction, which is cheap and stable but often leads
to complex deployment operators, whereas direct QAT optimizes the final task
objective but provides sparse and noisy supervision~\citep{frantar2023gptq,wu2024ptq4dit,chen2024qdit,jacob2018quantization,yang2025robuq}.
The intermediate region between these two extremes remains underexplored.

Based on this view, we propose \ours{}, a two-stage framework for training plain
4-bit-activation generative models.  Its core method, \method{}, starts from
local distillation objectives and progressively expands supervision to longer
low-bit computation paths before transitioning to standard end-to-end QAT or
QAD.  This moves the required complexity into training while leaving deployment
as a plain low-bit operator graph.

\begin{figure*}[t]
      \vspace{-10mm}
      \centering
      \begin{minipage}{0.49\textwidth}
          \centering
          \IfFileExists{figures/holder/5_png/ours_1.png}{\includegraphics[width=0.188\linewidth]{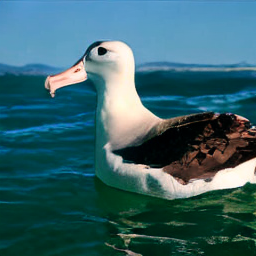}}{\fbox{\rule{0pt}{0.18\linewidth}\rule{0.16\linewidth}{0pt}}}\hspace{0.2mm}%
          \IfFileExists{figures/holder/5_png/ours_2.png}{\includegraphics[width=0.188\linewidth]{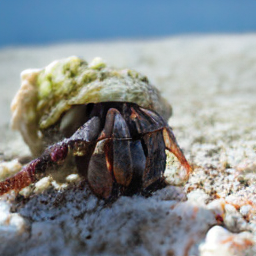}}{\fbox{\rule{0pt}{0.18\linewidth}\rule{0.16\linewidth}{0pt}}}\hspace{0.2mm}%
          \IfFileExists{figures/holder/5_png/ours_3.png}{\includegraphics[width=0.188\linewidth]{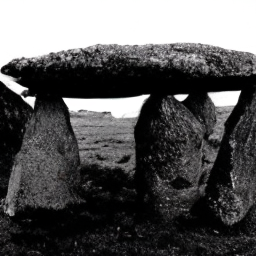}}{\fbox{\rule{0pt}{0.18\linewidth}\rule{0.16\linewidth}{0pt}}}\hspace{0.2mm}%
          \IfFileExists{figures/holder/5_png/ours_4.png}{\includegraphics[width=0.188\linewidth]{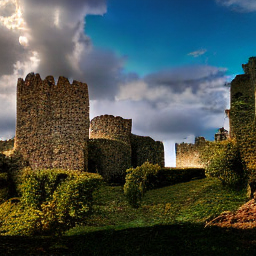}}{\fbox{\rule{0pt}{0.18\linewidth}\rule{0.16\linewidth}{0pt}}}\hspace{0.2mm}%
          \IfFileExists{figures/holder/5_png/ours_5.png}{\includegraphics[width=0.188\linewidth]{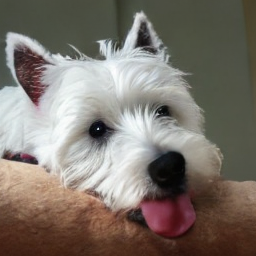}}{\fbox{\rule{0pt}{0.18\linewidth}\rule{0.16\linewidth}{0pt}}}\\[0.2mm]
          \IfFileExists{figures/holder/5_png/ours_11.png}{\includegraphics[width=0.188\linewidth]{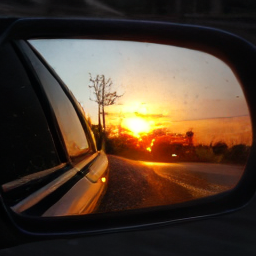}}{\fbox{\rule{0pt}{0.18\linewidth}\rule{0.16\linewidth}{0pt}}}\hspace{0.2mm}%
          \IfFileExists{figures/holder/5_png/ours_7.png}{\includegraphics[width=0.188\linewidth]{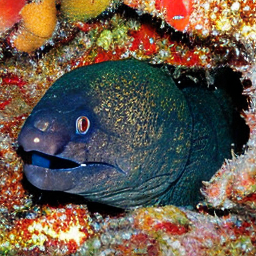}}{\fbox{\rule{0pt}{0.18\linewidth}\rule{0.16\linewidth}{0pt}}}\hspace{0.2mm}%
          \IfFileExists{figures/holder/5_png/ours_8.png}{\includegraphics[width=0.188\linewidth]{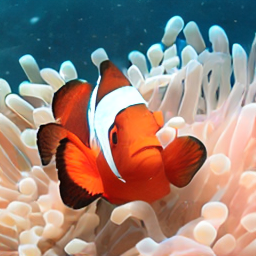}}{\fbox{\rule{0pt}{0.18\linewidth}\rule{0.16\linewidth}{0pt}}}\hspace{0.2mm}%
          \IfFileExists{figures/holder/5_png/ours_9.png}{\includegraphics[width=0.188\linewidth]{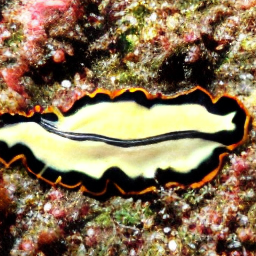}}{\fbox{\rule{0pt}{0.18\linewidth}\rule{0.16\linewidth}{0pt}}}\hspace{0.2mm}%
          \IfFileExists{figures/holder/5_png/ours_10.png}{\includegraphics[width=0.188\linewidth]{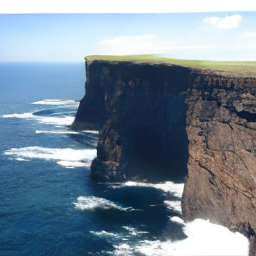}}{\fbox{\rule{0pt}{0.178\linewidth}\rule{0.166\linewidth}{0pt}}}
      \end{minipage}
      \hfill
      \begin{minipage}{0.49\textwidth}
          \centering
          \begin{minipage}{0.48\linewidth}
              \centering
              \includegraphics[height=0.9\textwidth,keepaspectratio]{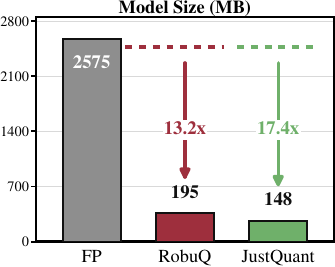}
          \end{minipage}
          \hfill
          \begin{minipage}{0.48\linewidth}
              \centering
              \includegraphics[height=0.9\textwidth,keepaspectratio]{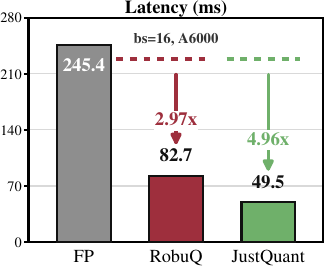}
          \end{minipage}
      \end{minipage}
      \vspace{-3mm}
      \caption{\textbf{Naive W1.58A4 ImageNet samples and deployment summary.} With a 24-hour GPT5.6-sol kernel-optimization budget, the clean \ours{} inference path realizes latency gains more easily than operator-heavy alternatives, even on a 0.6B DiT where small matrices and non-GEMM time make compute-bound speedup difficult.}
      \label{fig:w1p58a4_samples}
      \vspace{-5mm}
  \end{figure*}

We evaluate \ours{} in controlled and data-limited regimes.  Our main
experiments use DiT-XL/2 to study the optimization behavior of plain A4
quantization, and we further extend to larger text-to-image DiT and dLLM models
under limited-budget and limited-data settings.
Figure~\ref{fig:w1p58a4_samples} gives a visual and deployment summary of the
naive W1.58A4 setting.

Our main contributions are summarized as follows:
\begin{itemize}
    \item \textbf{A distillation view of plain A4 quantization.} We frame
    4-bit activation quantization as a Net2Net knowledge distillation problem
    for plain low-bit graphs.  We analyze why PTQ and QAT methods struggle:
    PTQ often depends on complex deployment-side operators, while direct QAT
    suffers from stacked STE approximations and sparse supervision.
    \item \textbf{A progressive multi-level distillation algorithm.} We propose
    \ours{} and its core method, \method{}, which uses progressive supervision
    levels as a warm-up before standard end-to-end QAT or QAD.  The method moves
    the missing complexity into training while keeping the deployed model as a
    plain low-bit operator graph.
    \item \textbf{Controlled small-model regime.} On DiT-XL/2, \method{} brings
    plain A4 quantization back to frontier-level generation quality.  In the
    extreme W1.58A4 setting, it reduces FID-50K from 5.5698 to 3.3021 and raises
    IS-50K from 166.06 to 241.95.  We further find that the progressive warm-up
    unexpectedly preserves generative diversity, addressing a previously
    unresolved failure mode of direct low-bit QAT.
    \item \textbf{Data-limited large-model regime.} We extend the same principle
    to larger, data-limited settings, including large text-to-image DiT and dLLM
    models.  \method{} outperforms PTQ and QAD
    baselines without adding new operators to the inference graph.
\end{itemize}

\vspace{-4mm}
\section{Background and Related Work}
\label{sec:related_work}

\paragraph{Specialized operators in PTQ and QAT.}
A major line of PTQ and QAT improves low-bit quality by changing the computation
executed at inference.  PTQ methods smooth or rotate activations, permute
channels, factorize weights, or preserve mixed-precision branches to reduce
outliers and reconstruction error~\citep{xiao2023smoothquant,wei2022qdrop,ashkboos2024quarot,liu2024spinquant,lin2025duquant,li2024svdquant,shao2025blockrotation}.
For LLMs, BWLA uses orthogonal transforms and low-rank refinement for binary
weights with low-bit activations, while SpecQuant combines smoothing with
frequency-domain truncation~\citep{zhao2026bwla,zhao2026specquant}.
Other ultra-low-bit PTQ approaches separate shared and expert-specific MoE
weight components for vector quantization or use binary factorization under
a strict bit budget
\citep{xu2026kbvq,zhao2026af1}.
Other methods modify the operator itself: Quantizable Transformers changes
attention to avoid learned outliers, SVDQuant and Q-DiT4SR add low-rank
branches, and MXFP4/NVFP4-style work relies on specialized numerical
formats~\citep{bondarenko2023quantizable,li2024svdquant,zhang2026qdit4sr,xin2026qadnvfp4,fourtune2026}.  QAT methods follow the same pattern through
Hadamard transforms, robust activation quantizers, modified normalization,
rotation-based modules, or attention-specific kernels
\citep{wang2025bitnetv2,panferov2025quest,yang2025robuq,huang2025convrot,xu2025vetadit,cheng2026permuquant,zhang2026attnqat}.  These techniques can
recover accuracy, but their benefits come with extra transforms, branches,
kernels, or format-specific hardware.
Hardware-oriented designs such as QUARK additionally accelerate quantized
Transformer nonlinearities through FPGA circuit sharing~\citep{zhao2025quark}.
We study the complementary setting:
deployment remains a naive low-bit operator graph, and the missing complexity
is handled during training.

\paragraph{The evolution of DiT and dLLM quantization.}
Quantization for generative transformers has progressed from weight-only
compression toward joint weight--activation quantization.  Early diffusion
work identifies two recurring obstacles: activation distributions shift with
timestep, and quantization error accumulates across denoising
steps~\citep{li2023qdm,he2023ptqd,he2024efficientdm}.  DiT-specific PTQ then
adds channel salience balancing, timestep-aware calibration, and token-aware
granularity~\citep{wu2024ptq4dit,chen2024qdit,zhao2024viditq}.  Later methods
push toward more aggressive 4-bit or sub-4-bit settings with low-rank
branches, rotations, temporal adaptation, binarization, and mixed-precision
search~\citep{li2024svdquant,ryu2025dgq,zheng2024bidm,zheng2025binarydm,huang2025convrot,xu2025vetadit,yang2025treeq,zhang2026qdit4sr}.
Training-based systems such as TerDiT and RobuQ further show that ternary or
sub-4-bit DiT can be optimized, but still rely on specialized inference
paths~\citep{lu2024terdit,yang2025robuq}.  Diffusion language models expose the
same pattern in text generation: their masked-denoising activations and dynamic
mask ratios make ordinary autoregressive-LLM calibration unreliable, motivating
dLLM-specific PTQ methods~\citep{zhang2026quantdllm,chen2026dllmquant}.  These
developments leave open whether a plain W4A4 or W1.58A4 graph can be reached
reliably without adding deployment operators.

\paragraph{Distillation for quantized models.}
Knowledge distillation transfers the behavior of a high-precision teacher to a
compact student~\citep{hinton2015distilling}.  In quantization, the teacher can
supervise logits, attention maps, hidden states, local blocks, or generative
trajectories.  Q-ViT uses distribution-guided distillation to repair attention
distortion in fully quantized vision transformers, while token-scaled logit
distillation adjusts the logit signal for ternary-weight generative language
models~\citep{li2022qvit,kim2023tokenscaled}.  Other work uses hidden-state or
block-wise objectives to make compressed students easier to train
\citep{dasgupta2025hiddenstate,chen2024efficientqat}.  In diffusion models,
BiDM, BinaryDM, and Q-VDiT use spatial, low-rank, or temporal distillation to
protect generative structure under aggressive quantization
\citep{zheng2024bidm,zheng2025binarydm,feng2025qvdit}.
DilateQuant combines block-wise knowledge distillation (BKD) with Weight
Dilation and a Temporal Parallel Quantizer~\citep{liu2024dilatequant}.
EfficientQAT trains all parameters block-wise before end-to-end training of
quantization step sizes~\citep{chen2024efficientqat}. Our contribution is not
block-wise alignment itself, but progressively merging teacher-fed student
segments and allocating training steps across their lengths, while retaining
plain deployed operators. This complements output-level QAD
\citep{xin2026qadnvfp4} and capacity-aware supervision
\citep{harutyunyan2023supervision}.

\vspace{-4mm}
\section{Method}
\label{sec:method}
  \begin{figure*}[t]
      \vspace{-10mm}
      \centering
      \includegraphics[width=1.0\textwidth]{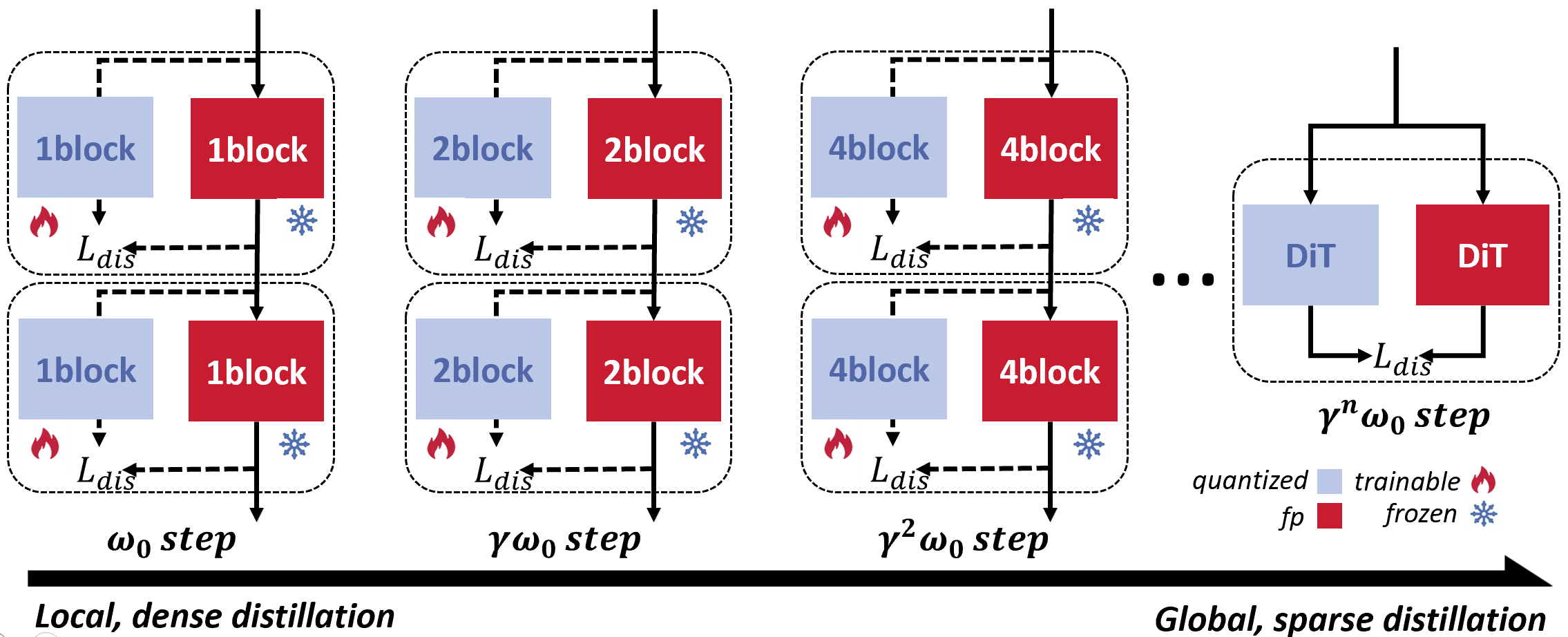}
      \vspace{-8mm}
      \caption{\textbf{Theseus QAD.} Theseus QAD expands distillation from local blocks to the full model while allocating training steps geometrically across path lengths.}
      \label{fig:overview}
      \vspace{-7mm}
  \end{figure*}
\vspace{-2mm}
\subsection{Quantization as distillation}
\label{sec:formulation}

\begin{wrapfigure}{r}{0.5\textwidth}
    \vspace{-10mm}
    \centering
    \includegraphics[width=0.5\textwidth]{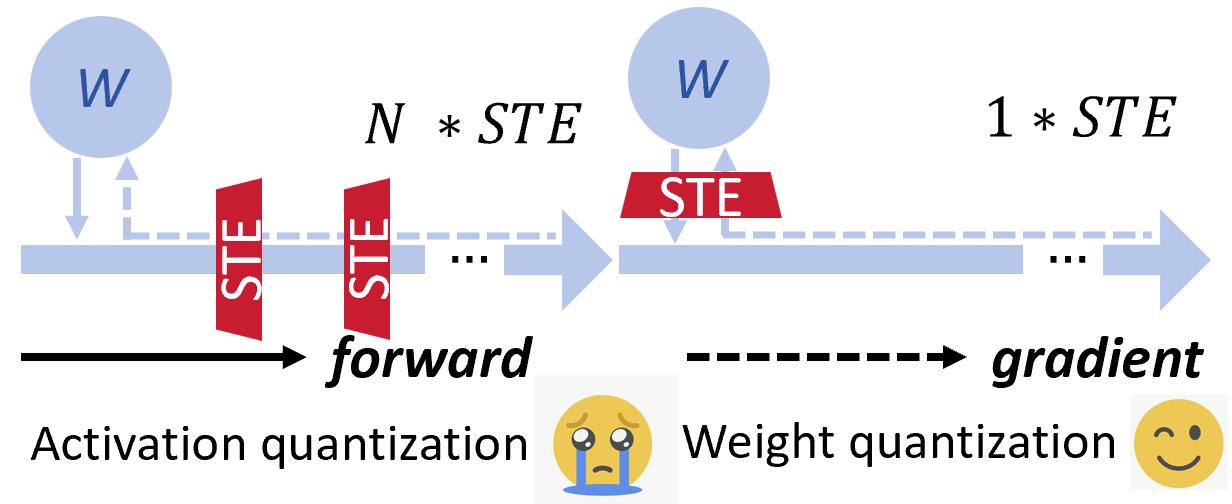}
    \vspace{-8mm}
    \caption{\textbf{Why activation quantization is hard to QAT.} }
    \label{fig:ste_instability}
    \vspace{-5mm}
\end{wrapfigure}
\vspace{-2mm}
We view quantization as a Net2Net-style replacement guided by knowledge
distillation.  Let the full-precision network be
\[
    f_{\theta}(x)=B_L \circ B_{L-1}\circ \cdots \circ B_1(x),
\]
where each $B_i$ is a replaceable block or sub-block.  Its quantized counterpart
preserves the topology,
\[
    q_{\phi}(x)=\widetilde{B}_L \circ \widetilde{B}_{L-1}\circ \cdots \circ
    \widetilde{B}_1(x),
\]
with each module replaced by a low-bit operator from the target deployment
family.  The goal is therefore to preserve network behavior, rather than merely
reconstruct isolated matrices. This view separates the deployed operator from the training path: \method{}
keeps the deployed A4 operator plain and moves the additional guidance into
training.

\textbf{Why PTQ fails.} PTQ mainly focuses on local linear layers or short
blocks by matching their outputs to the corresponding full-precision
activations.  This signal is dense and cheap, but local agreement does not
ensure that multiple replacements compose into a good network.  Aggressive PTQ
therefore often introduces smoothing, rotations, Hadamard transforms, SVD or
low-rank branches, mixed precision, which improve local
approximation at the cost of a more complex deployed operator.

\textbf{Why QAT fails.} QAT optimizes the deployed low-bit model with the final task. Although aligned with final model, its correction signal is sparse and delayed. This is damaging for 4-bit activations, where the hidden state passes through straight-through estimator (STE) approximations, causing biased and noisy gradients (Figure~\ref{fig:ste_instability}). Quantization-aware distillation (QAD) adds a teacher, but forms still use one fixed alignment level: output-only QAD matches final behavior, while local QAD matches short segments. Neither setting exploits supervision at multiple  granularities.

\subsection{Distillation granularity and scheduling}
\label{sec:granularity_tradeoff}

We do not adopt a single fixed distillation width as a general design:
a width suited to one model need not transfer to another architecture or
quantization setting. Instead, we schedule distillation granularity according
to three properties:

\noindent
\begin{minipage}[t]{0.31\linewidth}
\small
\textbf{1. Training stability.} Activation quantization makes optimization
intrinsically unstable.  Longer low-bit paths contain more stacked
straight-through estimator (STE) approximations, so starting with end-to-end
distillation can easily cause training to fail.
\end{minipage}\hfill
\begin{minipage}[t]{0.33\linewidth}
\small
\textbf{2. Model alignment.} Extremely low-bit entries cannot match their
full-precision counterparts exactly, while their composition can still
approximate the overall transformation.  An overly local target may therefore
over-constrain intermediate values that the quantized network does not need to
preserve.
\end{minipage}\hfill
\begin{minipage}[t]{0.31\linewidth}
\small
\textbf{3. Distillation density.} Lower-level targets expose more intermediate
states and provide denser supervision, allowing the student to receive useful
updates more quickly.  Global distillation provides the sparsest signal and is
therefore the slowest to optimize.
\end{minipage}

\vspace{-0.5mm}
These properties motivate progressive distillation granularity rather than
a single fixed target.  Our \emph{Ship of Theseus} analogy refers to merging
short low-bit segments into longer ones, not progressively replacing a mixed
teacher--student forward path.  Each level inherits the student parameters
from the preceding level.  Student blocks compose within each segment, but
segment inputs remain teacher activations; student outputs are not passed
between segments.  Full-model distillation connects the entire student path.
Subsequent QAT or QAD optimizes the model without changing its operators.

\subsection{Theseus QAD design: local-to-global distillation}
\label{sec:theseus_qad}
  \begin{figure*}[t]
      \vspace{-14mm}
      \centering
      \includegraphics[width=1.0\textwidth]{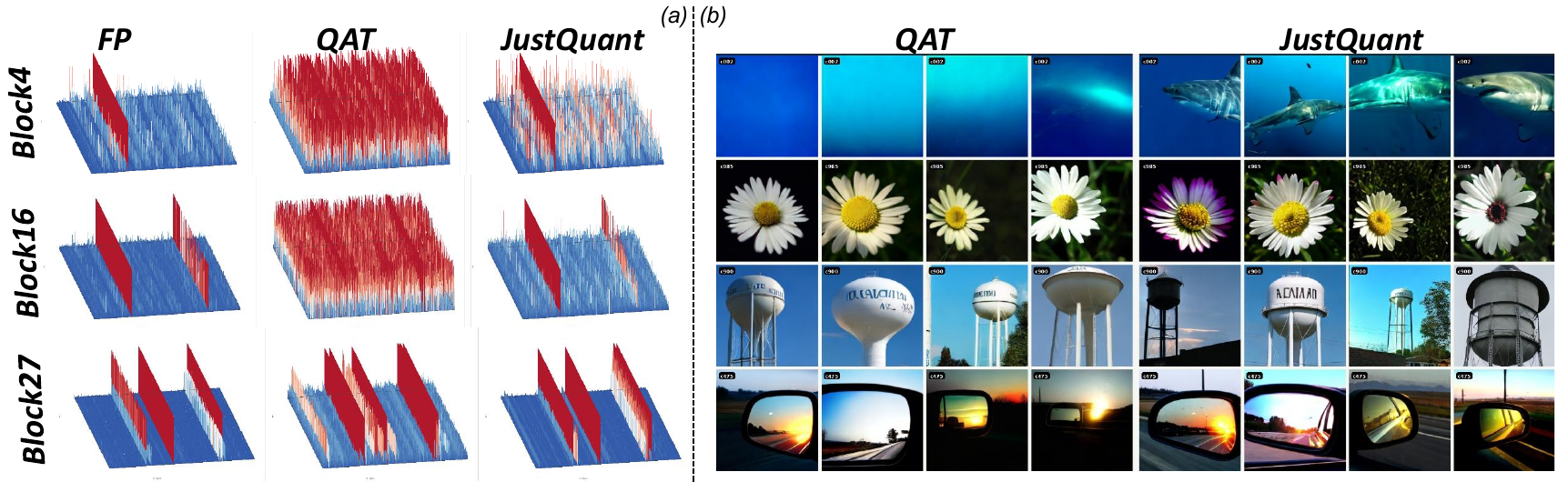}
      \vspace{-8mm}
      \caption{\textbf{Effect of Theseus QAD on W1.58A4 DiT.} (a) Intermediate activations after Theseus QAD warm-up and QAT resemble those of the FP model, whereas direct W1.58A4 QAT yields more disordered activations. (b) Theseus QAD substantially improves generation quality and diversity.}
      \label{fig:w158a4_activation}
      \vspace{-7mm}
  \end{figure*}
\method{} uses a local-to-global, teacher-forced distillation schedule,
illustrated in Figure~\ref{fig:overview}.
Figure~\ref{fig:w158a4_activation} provides a qualitative comparison on
W1.58A4 DiT: Theseus QAD warm-up followed by QAT yields intermediate
activations closer to the full-precision reference and better generation
quality than direct QAT.  Stage 1 trains the low-bit student by matching teacher
and student subnetworks at multiple depths.  For a group of width $w$ starting
at block $g$, both networks receive the same teacher state $h_g$:
\[
\begin{aligned}
    t_{g,w} &= B_{g+w-1}\circ \cdots \circ B_g(h_g),\\
    s_{g,w} &= \widetilde{B}_{g+w-1}\circ \cdots \circ
              \widetilde{B}_g(h_g).
\end{aligned}
\]
Stage 1 minimizes
\[
    \mathcal{L}_{\mathrm{S1}}
    = \mathbb{E}_{x,t,g,w}\left[d(s_{g,w},t_{g,w})\right],
\]
where $t$ denotes task-specific conditioning, such as a diffusion timestep.  In
our DiT experiments, $d$ is an activation reconstruction loss and the default
implementation uses $L_1$ distance.

All distillation paths use the same binary-merge schedule: starting from
one-block segments, we repeatedly merge neighboring segments, giving the
fixed widths $\mathcal W=(1,2,4,8,16,28)$ for the 28-block DiT.  The parameter
$\gamma$ does not define these widths.  It only controls the geometric
allocation of the Stage-1 budget across this fixed schedule:
\[
    N_k\propto\gamma^k,\qquad \sum_{k=0}^{K}N_k=T_1,
    \qquad 0<\gamma<\infty.
\]
Here $k$ indexes the fixed widths from shortest to longest. Integer step
allocations sum to the prescribed budget; the exact allocations used in the
DiT sweep are listed in Figure~\ref{fig:gamma_sweep}.
Thus, $\gamma=1$ gives a uniform allocation, while larger $\gamma$ values
emphasize longer student segments; the $\gamma=0$ and $\gamma=\infty$ rows
are the corresponding shortest-path and longest-path endpoint schedules.
For these endpoints, all $T_1$ steps are assigned to $w_0$ or $w_K$,
respectively, and zero-budget levels are skipped.
Thus, $\gamma=0$ uses only single-block distillation in Stage 1, reflecting
the local alignment principle in DilateQuant and EfficientQAT
\citep{liu2024dilatequant,chen2024efficientqat}. Conversely, $\gamma=\infty$ denotes full-model QAD~\citep{xin2026qadnvfp4}, with all Stage-1 steps assigned to the longest path.

The sweep in Fig.~\ref{fig:gamma_sweep} therefore tests our central hypothesis:
the appropriate distillation signal should be scheduled across granularities,
rather than selected once and held fixed.  The two single-granularity extremes
are both suboptimal: using only 1-block supervision ($\gamma=0$) is too local,
while jumping directly to full-model distillation ($\gamma=\infty$) is too
unstable.  This explains why prior fixed-granularity distillation can fail.
In contrast, A4 targets benefit from allocating more exposure to wider
distillation groups, whereas the W1.58A32 weight-only reference prefers a more
balanced schedule. Reversing the order raises FID from 6.28 to 8.70 for
W4A4 ($\gamma=2$) and from 12.87 to 29.50 for W1.58A32 ($\gamma=1$),
supporting short-to-long distillation in these settings. Appendix~\ref{app:faq} addresses common questions about our
  design choices, evaluation, and deployment scope.

\begin{figure*}[t]
\vspace{-10mm}
\centering
\newlength{\stagefigheight}
\setlength{\stagefigheight}{0.315\textwidth}
\begin{minipage}[t]{0.48\textwidth}
\vspace{0pt}
\begin{minipage}[t][\stagefigheight][t]{\linewidth}
\scriptsize
\hrule height 0.5pt
\vspace{0.7mm}
\textbf{Algorithm 1} Theseus QAD: local-to-global distillation
\vspace{0.6mm}
\hrule height 0.35pt
\vspace{0.8mm}
\textbf{func} \textsc{StageOne}($f_{\theta},q_{\phi},\mathcal{D},\gamma,T_1$)\\
\textbf{Input:} teacher $f_{\theta}$, student $q_{\phi}$, data $\mathcal{D}$, allocation ratio $\gamma$, budget $T_1$\\[-0.2mm]
\begin{tabular}{@{}r@{\hspace{0.6mm}}p{0.86\linewidth}@{}}
1: & $\phi\leftarrow\theta,\quad (w_0,\ldots,w_K)\leftarrow\textsc{BinaryMergeWidths}(L)$ \\
2: & $(N_0,\ldots,N_K)\leftarrow\textsc{Allocate}(T_1,\gamma)$ (see text) \\
3: & \textbf{for} $k=0,\ldots,K$ and $u=1,\ldots,N_k$ \textbf{do} \\
4: & \quad $(x,t)\sim\mathcal{D}$,\quad $h\leftarrow x,\quad \mathcal{L}_{\mathrm{S1}}\leftarrow0$ \\
5: & \quad \textbf{for} $g=0,w_k,2w_k,\ldots,L-w_k$ \textbf{do} \\
6: & \quad\quad $a\leftarrow h,\quad h\leftarrow B_{g+w_k-1}\circ\cdots\circ B_g(a)$ \\
7: & \quad\quad $s\leftarrow\widetilde{B}_{g+w_k-1}\circ\cdots\circ\widetilde{B}_g(a)$ \\
8: & \quad\quad $\mathcal{L}_{\mathrm{S1}}\leftarrow\mathcal{L}_{\mathrm{S1}}+d(s,h)$ \\
9: & \quad $\phi\leftarrow\phi-\eta\nabla_{\phi}\mathcal{L}_{\mathrm{S1}}$ \\
10: & \textbf{return} $q_{\phi}^{(1)}\leftarrow q_{\phi}$ \\
\end{tabular}
\vfill
\hrule height 0.5pt
\end{minipage}
\end{minipage}
\hfill
\begin{minipage}[t]{0.48\textwidth}
\vspace{0pt}
\begin{minipage}[t][\stagefigheight][t]{\linewidth}
\scriptsize
\hrule height 0.5pt
\vspace{0.7mm}
\centering
\renewcommand{\arraystretch}{1.12}
\setlength{\tabcolsep}{0.55mm}
\newcommand{\nstep}[1]{\makebox[2.35em][r]{#1}}
\begin{tabular*}{\linewidth}{@{\extracolsep{\fill}}llcccc@{}}
\multicolumn{2}{l}{\textbf{Table 1} Stage-1 ratio sweep} & \multicolumn{2}{c}{W4A4} & \multicolumn{2}{c}{W1.58A32} \\
\cmidrule(lr){3-4} \cmidrule(lr){5-6}
$\gamma$  & $N_{1,2,4,8,16,28}$ & FID$\downarrow$ & IS$\uparrow$ & FID$\downarrow$ & IS$\uparrow$ \\
\midrule
0 & (\nstep{10000},\nstep{0},\nstep{0},\nstep{0},\nstep{0},\nstep{0}) & 8.11 & 177 & 35.42 & 63 \\
0.1 & (\nstep{9000},\nstep{900},\nstep{90},\nstep{9},\nstep{1},\nstep{0}) & 18.89 & 98 & 48.39 & 45 \\
0.33 & (\nstep{6676},\nstep{2225},\nstep{742},\nstep{247},\nstep{82},\nstep{28}) & 14.20 & 124 & 23.00 & 91 \\
0.5 & (\nstep{5079},\nstep{2540},\nstep{1270},\nstep{635},\nstep{317},\nstep{159}) & 9.23 & 162 & 17.16 & 113 \\
1 & (\nstep{1667},\nstep{1667},\nstep{1667},\nstep{1667},\nstep{1666},\nstep{1666}) & 6.58 & 223 & \best{12.87} & \best{146} \\
2 & (\nstep{159},\nstep{317},\nstep{635},\nstep{1270},\nstep{2540},\nstep{5079}) & \best{6.28} & \best{225} & 14.50 & 132 \\
3 & (\nstep{28},\nstep{82},\nstep{247},\nstep{742},\nstep{2225},\nstep{6676}) & 7.32 & 213 & 14.83 & 127 \\
10 & (\nstep{0},\nstep{1},\nstep{9},\nstep{90},\nstep{900},\nstep{9000}) & 6.96 & 218 & 15.69 & 113 \\
$\infty$ & (\nstep{0},\nstep{0},\nstep{0},\nstep{0},\nstep{0},\nstep{10000}) & 188.26 & 4 & 18.05 & 103 \\
\midrule
\multicolumn{2}{l}{Rev. ($\gamma_{\mathrm{W4A4}}=2,\;\gamma_{\mathrm{W1.58A32}}=1$)} & 8.70 & 169 & 29.50 & 82 \\
\end{tabular*}
\vfill
\hrule height 0.5pt
\end{minipage}
\end{minipage}
\vspace{-2mm}
\caption{\textbf{Theseus QAD warm-up and Stage-1 allocation.} The left panel shows the online Stage-1 loop: during one FP forward, each reached segment is immediately matched by its quantized counterpart, and the student is updated once after the accumulated loss. The right panel reports the same sweep, allocating 10k Stage-1 steps across replacement widths $(1,2,4,8,16,28)$, followed by 10k-step official QAT.}
\label{fig:gamma_sweep}
\vspace{-7.5mm}
\end{figure*}

\vspace{-2mm}
\section{Experiments}
\label{sec:experiments}
\vspace{-1mm}
\subsection{Setup}
\label{sec:exp_setup}

\begin{wraptable}{r}{0.4\columnwidth}
\vspace{-8mm}
\caption{\textbf{Experimental summary.}}
\label{tab:experimental_matrix}
\centering
\scriptsize
\setlength{\tabcolsep}{2.0mm}
\renewcommand{\arraystretch}{1.12}
\resizebox{\linewidth}{!}{%
\begin{tabular}{@{}lcc@{}}
\toprule[0.15em]
 & Small model & Large model \\
\midrule
DiT
& \begin{tabular}{c}DiT-XL/2, 0.6B\\[0.3mm]\citep{peebles2023dit}\end{tabular}
& \begin{tabular}{c}FLUX.1, 12B\\[0.3mm]\citep{blackforestlabs2024flux}\end{tabular} \\
\addlinespace
DLLM
& \begin{tabular}{c}ELF-B, 105M\\[0.3mm]\citep{hu2026elf}\end{tabular}
& \begin{tabular}{c}LLaDA, 8B\\[0.3mm]\citep{nie2025llada}\end{tabular} \\
\bottomrule[0.15em]
\end{tabular}
}
\vspace{-7mm}
\end{wraptable}

DiT is the primary testbed for validation and ablation; DLLM experiments test
transfer across tasks.

\textbf{Datasets and evaluation metrics.}
We evaluate \ours{} in two regimes.  The first uses class-conditional
DiT-XL/2~\citep{peebles2023dit} on ImageNet~\citep{russakovsky2015imagenet}
$256\times256$, where data and compute are controllable enough to study plain
A4 operators.  FID-10K uses 10,000 generated images, CFG~\citep{ho2022cfg}
1.5, and 50 sampling steps; FID-50K uses 50,000 images, CFG 1.5, and 250
steps.  We report FID~\citep{heusel2017gans}, sFID
\citep{salimans2016improved,nash2021generating}, and IS
\citep{salimans2016improved,barratt2018anote} under the ADM protocol
\citep{dhariwal2021diffusion}.  The second uses FLUX.1-schnell and
FLUX.1-dev~\citep{blackforestlabs2024flux}, where full low-bit retraining is
expensive and the goal is recovery from limited data.  We report MJHQ FID,
CLIP-IQA~\citep{wang2022exploring}, CLIPScore~\citep{hessel2021clipscore},
ImageReward~\citep{xu2023imagereward}, and GenEval
\citep{ghosh2023geneval}.

\textbf{Compared methods.}
We compare against full-precision references, direct QAT and QAD using the
same plain operator family, and PTQ methods that introduce smoothing, rotations,
SVD branches, or mixed precision: SVDQuant~\citep{li2024svdquant},
ConvRot~\citep{huang2025convrot}, QuaRot~\citep{ashkboos2024quarot},
VETA-DiT~\citep{xu2025vetadit}, and TreeQ~\citep{yang2025treeq}.  We also
include trained low-bit systems such as TerDiT~\citep{lu2024terdit} and
RobuQ~\citep{yang2025robuq}, which add normalization, rotation, or SVD-based
components.  The tables identify these additional operators.

\textbf{Training and quantization details.}
For small models, we use Theseus QAD followed by task-level QAT.  For large
models, we use Theseus QAD alone, because end-to-end QAT becomes unstable at
the small batch sizes required by these experiments.  A representative
20k-step W4A4 DiT-XL/2 run takes about 1.75 GPU-hours on four GPUs, while the
FLUX experiments take about 0.75 days on four H200 GPUs.  Further training and
quantization details are provided in the appendix.

\vspace{-3mm}
\subsection{Main Results on Diffusion Models}
\label{sec:exp_diffusion}

\begin{table*}[!t]
\vspace{-14mm}
\caption{\textbf{DiT-XL/2 on ImageNet $256\times256$.} Further evaluation details are provided in Appendix~\ref{app:experimental_details}.}
\vspace{-3mm}
\label{tab:small_dit_main}
\begin{center}
\scriptsize
\setlength{\tabcolsep}{2.3mm}
\resizebox{\textwidth}{!}{
\begin{tabular}{clclcccl}
\toprule[0.15em]
Setting & Method & W/A & Type/Step & FID$\downarrow$ & sFID$\downarrow$ & IS$\uparrow$ & Additional Operator \\
\midrule
\multirow{15}{*}{\begin{tabular}{@{}c@{}}10K samples\\50 steps\end{tabular}} & FP~\citep{peebles2023dit} & 32/32 & -/7000k & 6.78 & 20.56 & 243.70 & -- \\
\cmidrule(lr){2-8}
& SVDQuant~\citep{li2024svdquant} & 4/4 & PTQ/0k & 81.28 & 67.27 & 29.17 & \textcolor{red!70!black}{Smooth, SVD} \\
& ConvRot~\citep{huang2025convrot} & 4/4 & PTQ/0k & 18.69 & 32.06 & 139.54 & \textcolor{red!70!black}{Rotation} \\
& QuaRot~\citep{ashkboos2024quarot} & 4/4 & PTQ/0k & 53.31 & 56.74 & 53.12 & \textcolor{red!70!black}{Rotation} \\
& VETA-DiT~\citep{xu2025vetadit} & 4/4 & PTQ/14.4k & 9.87 & 25.52 & 202.55 & \textcolor{red!70!black}{Smooth, Rotation} \\
& TreeQ~\citep{yang2025treeq} & 4/4 & PTQ/36.9k & 6.92 & 20.86 & 219.66 & \textcolor{red!70!black}{MP, SVD, Rotation} \\
& RobuQ~\citep{yang2025robuq} & 4/4 & QAT/20k & 8.20 & 22.94 & 222.02 & \textcolor{red!70!black}{Norm, SVD, Rotation} \\
& QAD~\citep{xin2026qadnvfp4} & 4/4 & QAD/20k & 188.26 & 190.47 & 3.89 & \textcolor{green!45!black}{None} \\
& QAT~\citep{jacob2018quantization} & 4/4 & QAT/20k & 10.72 & 26.55 & 196.53 & \textcolor{green!45!black}{None} \\
& \ours{} & 4/4 & QAT/20k & \best{6.48} & \best{19.31} & \best{238.31} & \textcolor{green!45!black}{None} \\
\cmidrule(lr){2-8}
& TerDiT~\citep{lu2024terdit} & 1.58/32 & QAT/1750k & 8.23 & 20.57 & 172.23 & \textcolor{red!70!black}{Norm} \\
& \ours{} & 1.58/32 & QAT/110k & \best{6.56} & \best{19.84} & \best{249.65} & \textcolor{green!45!black}{None} \\
\cmidrule(lr){2-8}
& RobuQ~\citep{yang2025robuq} & 1.58/4 & QAT/100k & 7.76 & 20.81 & 195.98 & \textcolor{red!70!black}{Norm, SVD, Rotation} \\
& QAT~\citep{jacob2018quantization} & 1.58/4 & QAT/100k & 9.93 & 20.33 & 155.85 & \textcolor{green!45!black}{None} \\
& \ours{} & 1.58/4 & QAT/110k & \best{7.23} & \best{19.88} & \best{223.99} & \textcolor{green!45!black}{None} \\
\midrule
\multirow{12}{*}{\begin{tabular}{@{}c@{}}50K samples\\250 steps\end{tabular}} & FP~\citep{peebles2023dit} & 32/32 & -/7000k & 2.27 & 4.55 & 277.83 & -- \\
\cmidrule(lr){2-8}
& SVDQuant~\citep{li2024svdquant} & 4/4 & PTQ/0k & 60.01 & 37.85 & 39.26 & \textcolor{red!70!black}{Smooth, SVD} \\
& ConvRot~\citep{huang2025convrot} & 4/4 & PTQ/0k & 7.98 & 13.36 & 187.96 & \textcolor{red!70!black}{Rotation} \\
& VETA-DiT~\citep{xu2025vetadit} & 4/4 & PTQ/14.4k & 2.99 & 6.27 & 247.05 & \textcolor{red!70!black}{Smooth, Rotation} \\
& RobuQ~\citep{yang2025robuq} & 4/4 & QAT/20k & \best{2.71} & \best{5.20} & \best{265.02} & \textcolor{red!70!black}{Norm, SVD, Rotation} \\
& QAT~\citep{jacob2018quantization} & 4/4 & QAT/20k & 5.15 & 8.34 & 220.53 & \textcolor{green!45!black}{None} \\
& \ours{} & 4/4 & QAT/20k & 3.32 & 5.79 & 256.56 & \textcolor{green!45!black}{None} \\
\cmidrule(lr){2-8}
& TerDiT~\citep{lu2024terdit} & 1.58/32 & QAT/1750k & 4.34 & 4.99 & 183.49 & \textcolor{red!70!black}{Norm} \\
& \ours{} & 1.58/32 & QAT/110k & \best{3.05} & \best{4.59} & \best{266.83} & \textcolor{green!45!black}{None} \\
\cmidrule(lr){2-8}
& RobuQ~\citep{yang2025robuq} & 1.58/4 & QAT/100k & \best{3.21} & 4.86 & 217.04 & \textcolor{red!70!black}{Norm, SVD, Rotation} \\
& QAT~\citep{jacob2018quantization} & 1.58/4 & QAT/100k & 5.57 & 5.57 & 166.06 & \textcolor{green!45!black}{None} \\
& \ours{} & 1.58/4 & QAT/110k & 3.30 & \best{4.51} & \best{241.95} & \textcolor{green!45!black}{None} \\
\bottomrule[0.15em]
\end{tabular}
}
\end{center}
\vspace{-5mm}
\caption{\textbf{FLUX W4A4 block quantization on MJHQ and GenEval.}}
\vspace{-2mm}
\label{tab:flux_main}
\begin{center}
\small
\setlength{\tabcolsep}{1.3mm}
\resizebox{\linewidth}{!}{
\begin{tabular}{llcccccccc}
\toprule[0.15em]
\multicolumn{2}{c}{Setting} & \multicolumn{3}{c}{MJHQ} & \multicolumn{4}{c}{GenEval} & \\
\cmidrule(lr){1-2} \cmidrule(lr){3-5} \cmidrule(lr){6-9}
Model & Method & FID$\downarrow$ & CLIP-IQA$\uparrow$ & ImageReward$\uparrow$ & Single$\uparrow$ & Count$\uparrow$ & Color$\uparrow$ & Overall$\uparrow$ & Add. Op. \\
\midrule
\multirow{6}{*}{\begin{tabular}{@{}c@{}}FLUX.1-schnell\\(4 steps)\end{tabular}} & FP~\citep{blackforestlabs2024flux} & 19.15 & 0.936 & 0.959 & 0.9844 & 0.5938 & 0.7872 & 0.6677 & -- \\
& ConvRot~\citep{huang2025convrot} & \best{17.86} & 0.936 & 0.947 & 0.9812 & \best{0.6406} & 0.7819 & \best{0.6817} & \textcolor{red!70!black}{Rotation} \\
& SVDQuant~\citep{li2024svdquant} & 18.51 & \best{0.938} & \best{0.968} & \best{0.9906} & 0.6219 & \best{0.8138} & 0.6775 & \textcolor{red!70!black}{SVD} \\
\cdashline{2-10}
& Naive INT4 & 18.37 & 0.929 & 0.902 & 0.9844 & 0.5813 & 0.7846 & 0.6753 & \textcolor{green!45!black}{None} \\
& QAT~\citep{jacob2018quantization} & 17.98 & 0.925 & 0.912 & 0.9844 & 0.6625 & 0.7872 & 0.6825 & \textcolor{green!45!black}{None} \\
& JustQuant & \best{17.62} & \best{0.943} & \best{0.993} & \best{0.9938} & \best{0.6844} & \best{0.8351} & \best{0.7149} & \textcolor{green!45!black}{None} \\
\midrule
\multirow{7}{*}{\begin{tabular}{@{}c@{}}FLUX.1-dev\\(50 steps)\end{tabular}} & FP~\citep{blackforestlabs2024flux} & 20.26 & 0.952 & 0.958 & 0.9875 & \best{0.7281} & 0.7926 & \best{0.6707} & -- \\
& ConvRot~\citep{huang2025convrot} & 19.78 & \best{0.953} & 0.948 & 0.9844 & 0.7094 & 0.7979 & 0.6662 & \textcolor{red!70!black}{Rotation} \\
& SVDQuant~\citep{li2024svdquant} & \best{19.71} & 0.948 & \best{0.961} & \best{0.9969} & 0.6938 & \best{0.8032} & 0.6687 & \textcolor{red!70!black}{SVD} \\
\cdashline{2-10}
& Naive INT4 & 24.29 & 0.923 & 0.754 & 0.9844 & 0.7094 & 0.7952 & 0.6465 & \textcolor{green!45!black}{None} \\
& QAT~\citep{jacob2018quantization} & 22.54 & 0.917 & 0.520 & 0.9500 & 0.5312 & 0.6888 & 0.5302 & \textcolor{green!45!black}{None} \\
& QAD~\citep{xin2026qadnvfp4} & 20.44 & \best{0.955} & 0.898 & \best{0.9906} & 0.6906 & 0.7819 & 0.6585 & \textcolor{green!45!black}{None} \\
& JustQuant & \best{20.21} & 0.951 & \best{0.956} & \best{0.9906} & \best{0.7188} & \best{0.8032} & \best{0.6643} & \textcolor{green!45!black}{None} \\
\bottomrule[0.15em]
\end{tabular}
}
\end{center}
\vspace{-9mm}
\end{table*}
\vspace{-1mm}
\textbf{DiT.} Table~\ref{tab:small_dit_main} compares the 10K/50-step and
50K/250-step sampling protocols. Plain rows use low-bit operators without extra inference
transformations. In W4A4, direct QAD fails to recover the model (FID-10K
188.26), and direct QAT reaches FID-10K 10.72. With staged replacement followed
by QAT, \method{} reduces FID-10K to 6.48, close to the FP value of 6.78, and
improves IS from 196.53 to 238.31. The gain persists at 50K samples: FID drops
from 5.15 to 3.32, and IS rises from 220.53 to 256.56. For W1.58A4, where
both weight and activation precision are restricted, FID-50K improves from
5.57 to 3.30 and IS from 166.06 to 241.95 over direct QAT. The W1.58A32
result also improves on TerDiT despite using fewer training steps and no
additional normalization operator. Operator-heavy methods can still yield a
lower FID, as RobuQ does in W4A4. However, our method achieves greater sample diversity.

\textbf{FLUX.} Following the transformer-focused quantization practice of
SVDQuant~\citep{li2024svdquant}, we use W4A4 group64 for transformer-block
linear layers.
Table~\ref{tab:flux_main} reports MJHQ and representative GenEval
sub-scores; the full breakdown is in the appendix. Dashed lines distinguish
methods with extra inference operators from plain low-bit methods, and bold
marks the best result within each group. On FLUX.1-schnell, \method{} improves
MJHQ FID from 18.37 for naive INT4 to 17.62 and raises GenEval overall from
0.6753 to 0.7149, exceeding the tested rotation and SVD baselines on the
latter metric without their added operators. On FLUX.1-dev, direct QAT lowers
GenEval overall to 0.5302, while standard QAD recovers it to 0.6585.
\method{} reaches 0.6643 and improves MJHQ FID from QAD's 20.44 to 20.21;
its ImageReward (0.956) also approaches the FP value (0.958). QAD retains a
slight CLIP-IQA advantage, so the improvement is not uniform across metrics.
These results support staged distillation when full end-to-end QAT is unstable
at the batch sizes available for larger models.

\vspace{-4mm}
\subsection{Results on Diffusion Language Models}
\label{sec:exp_dllm}

\textbf{ELF}. ELF-B tests diffusion language generation under a larger training
budget.  The goal is to check whether \method{} still gives an advantage at
both W4A4 and W1.58A4 when the comparison moves beyond image generation.  We
compare against direct QAT and RobuQ under matched budgets. Table~\ref{tab:elf_owt} shows that \method{} is the strongest quantized method
for both bit widths.  In W4A4, its generation perplexity stays close to FP
under both Qwen3-8B-Base and GPT-2-Large judges, while RobuQ and direct QAT
introduce much larger gaps.  In W1.58A4, \method{} slightly improves over the
FP reference within the reported variance.  This suggests that, with enough
training, the staged objective remains effective for DLLM generation and does
not rely on DiTs.

\begin{table*}[t]
\vspace{-14mm}
\caption{\textbf{ELF-B OpenWebText generation.} W4A4 and W1.58A4 results are
reported with generation perplexity from two judge models, following the
evaluation protocol of ELF~\citep{hu2026elf}.}
\vspace{-3mm}
\label{tab:elf_owt}
\begin{center}
\scriptsize
\setlength{\tabcolsep}{2.0mm}
\resizebox{\textwidth}{!}{
\begin{tabular}{llccccc}
\toprule[0.15em]
\multirow{2}{*}{W/A} & \multirow{2}{*}{Arm} & \multicolumn{2}{c}{Qwen3-8B-Base} & \multicolumn{2}{c}{GPT-2-Large} & \multirow{2}{*}{Add. Op.} \\
\cmidrule(lr){3-4}\cmidrule(lr){5-6}
 &  & gen-PPL$\downarrow$(4seed) & $\Delta$PPL vs FP & gen-PPL$\downarrow$(4seed) & $\Delta$PPL vs FP & \\
\midrule
\multirow{4}{*}{4/4}
& FP & $14.729 \pm 0.152$ & -- & $19.029 \pm 0.305$ & -- & -- \\
& \ours{} & \best{$14.854 \pm 0.344$} & $+0.125$ & \best{$19.180 \pm 0.417$} & $+0.152$ & \textcolor{green!45!black}{None} \\
& RobuQ & $16.751 \pm 0.384$ & $+2.022$ & $22.089 \pm 0.529$ & $+3.061$ & \textcolor{red!70!black}{Hadamard, SVD} \\
& Direct QAT & $19.625 \pm 0.327$ & $+4.896$ & $25.011 \pm 0.465$ & $+5.983$ & \textcolor{green!45!black}{None} \\
\midrule
\multirow{4}{*}{1.58/4}
& FP & $14.823 \pm 0.208$ & -- & $19.254 \pm 0.253$ & -- & -- \\
& \ours{} & \best{$14.626 \pm 0.482$} & $-0.197$ & \best{$19.038 \pm 0.634$} & $-0.216$ & \textcolor{green!45!black}{None} \\
& RobuQ & $18.537 \pm 0.265$ & $+3.714$ & $25.707 \pm 0.414$ & $+6.453$ & \textcolor{red!70!black}{Hadamard, SVD} \\
& Direct QAT & $16.776 \pm 0.255$ & $+1.953$ & $22.051 \pm 0.391$ & $+2.797$ & \textcolor{green!45!black}{None} \\
\bottomrule[0.15em]
\end{tabular}
}
\end{center}
\vspace{-4mm}
\caption{\textbf{LLaDA-8B W4A4 on GSM8K and held-out transfer benchmarks.} }
\label{tab:llada_side_by_side}
\begin{center}
\scriptsize
\begin{minipage}[t][0.21\textwidth][t]{0.49\textwidth}
\centering
\textbf{(a) GSM8K metrics by method}
\vspace{1mm}
\setlength{\tabcolsep}{1.8mm}
\resizebox{\linewidth}{!}{
\begin{tabular}{lcccc}
\toprule[0.15em]
Method & Greedy$\uparrow$ & pass@1$\uparrow$ & pass@4$\uparrow$ & pass@16$\uparrow$ \\
\midrule
FP & \best{0.6748} & \best{0.6307} & \best{0.8539} & \best{0.9325} \\
PTQ & 0.6035 & 0.5704 & 0.8122 & 0.9113 \\
QAT & 0.6138 & 0.5836 & 0.8203 & 0.9163 \\
QAD & 0.6475 & 0.6015 & 0.8287 & 0.9196 \\
JustQuant & 0.6543 & 0.6137 & 0.8455 & 0.9249 \\
\bottomrule[0.15em]
\end{tabular}
}
\end{minipage}
\hfill
\begin{minipage}[t][0.20\textwidth][t]{0.49\textwidth}
\centering
\textbf{(b) Transfer after GSM8K-only distillation}
\vspace{1mm}
\setlength{\tabcolsep}{1.9mm}
\resizebox{\linewidth}{!}{
\begin{tabular}{lcccc}
\toprule[0.15em]
Method & PIQA & ARC-C & C-EVAL & HumanEval \\
\midrule
FP & 0.8300 & \best{0.8662} & \best{0.6440} & \best{0.4512} \\
PTQ & 0.8280 & 0.8495 & 0.6420 & 0.4268 \\
QAT & 0.8150 & 0.8200 & 0.6050 & 0.3512 \\
QAD & 0.8320 & 0.8562 & \best{0.6440} & \best{0.4512} \\
JustQuant & \best{0.8420} & 0.8562 & 0.6420 & 0.4451 \\
\bottomrule[0.15em]
\end{tabular}
}
\end{minipage}
\end{center}
\vspace{-13mm}
\end{table*}

\textbf{LLaDA.} We evaluate LLaDA-8B with NVFP4 W4A4, distilling only on GSM8K and testing both GSM8K and held-out transfer tasks. This experiment evaluates the transferability of \method{} to another quantization format, rather than independence from specialized numerical formats.
Table~\ref{tab:llada_side_by_side} shows that \method{} achieves the highest reported quantized GSM8K greedy accuracy, improving from 0.6035 for PTQ and 0.6475 for QAD to 0.6543, alongside the highest quantized pass@1 and pass@16. Transfer performance remains broadly comparable to QAD:\method{} improves PIQA, matches ARC-Challenge, stays within 0.002 of FP/QAD on C-EVAL, and remains close to QAD on HumanEval. Direct QAT, in contrast, reduces all four transfer scores relative to FP. These results suggest that \method{} can improve target-task recovery while largely preserving the cross-task performance of ordinary QAD.

\vspace{-3mm}
\subsection{Ablation Studies}
\label{sec:exp_ablation}

\begin{table*}[t]
\vspace{-14mm}
\caption{\textbf{Ablation Study.}}
\vspace{-4mm}
\label{tab:w4a4_ablation_summary}
\begin{center}
\scriptsize
\renewcommand{\arraystretch}{1.08}
\setlength{\tabcolsep}{2.0mm}
\begin{minipage}[t][53mm][t]{0.499\textwidth}
\centering
\textbf{(a) Learning-rate sensitivity.}
\label{tab:w4a4_lr_diagnostic}
\vspace{1mm}
\resizebox{\linewidth}{!}{
\begin{tabular}{lccccc}
\toprule[0.15em]
Method & LR & Steps & FID$\downarrow$ & sFID$\downarrow$ & IS$\uparrow$ \\
\midrule
QAT & 1e-4 & 20k & 24.60 & 28.39 & 90.01 \\
QAT & 1e-6 & 20k & 27.52 & 35.68 & 84.70 \\
QAT & 2e-6 & 20k & 25.77 & 36.19 & 89.09 \\
QAT & 5e-6 & 20k & 25.64 & 34.68 & 85.60 \\
QAT & 1e-5 & 20k & 24.94 & 33.47 & 88.06 \\
QAT & 2e-5 & 20k & 18.27 & 29.78 & 114.82 \\
QAT & 5e-5 & 20k & 10.72 & 26.55 & 196.53 \\
\midrule
\ours{} & 1e-4 & 20k & \best{6.61} & 18.83 & \best{243.60} \\
\ours{} & 5e-5 & 20k & 6.72 & \best{18.65} & 236.99 \\
\bottomrule[0.15em]
\end{tabular}
}
\end{minipage}
\hfill
\begin{minipage}[t][53mm][t]{0.49\textwidth}
\centering
\textbf{(b) Real-data allocation.}
\label{tab:w4a4_data_allocation}
\vspace{1mm}
\resizebox{\linewidth}{!}{
\begin{tabular}{llcccc}
\toprule[0.15em]
Stage-1 data & Stage-2 data & Best & FID$\downarrow$ & sFID$\downarrow$ & IS$\uparrow$ \\
\midrule
\multirow{2}{*}{1\%} & same 1\% & 5k & 15.12 & 23.76 & 351.71 \\
 & full 100\% & 10k & 6.90 & 18.66 & 238.20 \\
\multirow{2}{*}{5\%} & same 5\% & 5k & 7.83 & 19.53 & 282.55 \\
 & full 100\% & 5k & 6.59 & 18.72 & 231.79 \\
\multirow{2}{*}{10\%} & same 10\% & 5k & 6.93 & 19.43 & 249.19 \\
 & full 100\% & 5k & 6.39 & 18.78 & 227.25 \\
\multirow{2}{*}{20\%} & same 20\% & 5k & 6.55 & 18.42 & 246.24 \\
 & full 100\% & 5k & 6.94 & 18.86 & 216.13 \\
\multirow{2}{*}{50\%} & same 50\% & 5k & 6.61 & 18.93 & 224.64 \\
 & full 100\% & 5k & 6.59 & 18.71 & 226.93 \\
\midrule
100\% & 100\% & 5k & \best{6.28} & \best{18.32} & 225.03 \\
\bottomrule[0.15em]
\end{tabular}
}
\end{minipage}

\vspace{-8mm}
\setlength{\tabcolsep}{0.8mm}
\begin{minipage}[t]{0.31\textwidth}
\centering
\textbf{(c) Loss design.}
\label{tab:w4a4_loss_ablation}
\vspace{1mm}
\resizebox{\linewidth}{!}{
\begin{tabular}{llccc}
\toprule[0.15em]
Ablation & Loss variant & FID$\downarrow$ & sFID$\downarrow$ & IS$\uparrow$ \\
\midrule
\multirow{2}{*}{Stage-1} & L1 & \best{6.72} & 18.65 & \best{236.99} \\
 & MSE/L2 & 6.85 & \best{18.45} & 232.30 \\
\midrule
\multirow{3}{*}{Stage-2} & task only & 6.72 & 18.65 & \best{236.99} \\
 & convex mix & 6.62 & 18.78 & 218.25 \\
 & additive mix & \best{6.51} & \best{18.46} & 232.61 \\
\bottomrule[0.15em]
\end{tabular}
}
\end{minipage}
\hfill
\begin{minipage}[t]{0.32\textwidth}
\centering
\textbf{(d) Multi-stage Distilled W1.58A4 .}
\label{tab:w1p58a4_multistage}
\vspace{1mm}
\resizebox{\linewidth}{!}{
\begin{tabular}{lcccc}
\toprule[0.15em]
Teacher & $\gamma$ & FID$\downarrow$ & sFID$\downarrow$ & IS$\uparrow$ \\
\midrule
N/A & -- & 5.57 & 5.57 & 166.06 \\
FP& $1$ & 3.64 & 4.70 & 227.23 \\
W1.58A32& $1$ & 3.39 & 5.51 & \best{244.00} \\
W1.58A32& $2$ & \best{3.30} & \best{4.51} & 241.95 \\
\bottomrule[0.15em]
\end{tabular}
}
\end{minipage}
\hfill
\begin{minipage}[t]{0.345\textwidth}
\centering
\textbf{(e) RobuQ with Theseus QAD .}
\label{tab:robuq_training_path}
\vspace{1mm}
\resizebox{\linewidth}{!}{
\begin{tabular}{lcccc}
\toprule[0.15em]
Method & $\gamma$ & FID$\downarrow$ & sFID$\downarrow$ & IS$\uparrow$ \\
\midrule
W4A4, RobuQ  & -- & \best{2.71} & 5.20 & 265.02 \\
+ \method{} & 2 & 2.89 & \best{4.48} & \best{267.22} \\
\midrule
W1.58A4, RobuQ& -- & \best{3.21} & 4.86 & 217.04 \\
+ \method{} & 1 & 3.16 & \best{4.53} & \best{276.25} \\
\bottomrule[0.15em]
\end{tabular}
}
\end{minipage}
\vspace{-9mm}
\end{center}
\end{table*}
\vspace{-2.5mm}
\textbf{Learning-rate sensitivity.}
Direct W4A4 QAT is highly sensitive to the learning rate.  Its FID improves
from 24.60 at $10^{-4}$ to 10.72 at $5{\times}10^{-5}$, but this best tested
setting still trails \method{} by a wide margin: at the same learning rate,
\method{} reaches FID 6.72 versus 10.72 and IS 236.99 versus 196.53.  In
contrast, \method{} is comparatively stable across the two tested learning
rates, with FID 6.61 and 6.72.  Thus, the main improvement is not explained by
tuning.

\textbf{Real-data allocation.}
The data sweep separates the role of Stage 1 from that of the final task
objective.  When Stage 2 has access to the full dataset, using only 5\%-10\%
of the data in Stage 1 remains effective, reaching FID 6.59 and 6.39,
respectively.  By contrast, restricting both stages to the same 1\% subset
causes a large degradation to FID 15.12, while 5\%-20\% shared subsets are
usable but remain weaker than the full-data run.  These results show that
Stage 1 can operate with limited coverage, but the final task stage still
needs enough data to restore global behavior.

\textbf{Loss design.}
For Stage 1, $L_1$ gives better FID and IS than MSE, while MSE gives slightly better
sFID.  This trade-off is consistent with both losses providing a
local signal; the loss does not determine whether training
succeeds.  For Stage 2, the task-only objective gives the strongest IS, whereas
adding a distillation term slightly improves FID or sFID but lowers IS.  We
therefore keep Stage 2 task-only in the main experiments, allowing the final
stage to optimize the deployment task directly.

\textbf{Multi-stage distillation order.}
Using a W1.58A32 student as the Stage-1 teacher for W1.58A4 improves
FID-50K/IS-50K from 3.6419/227.23 to 3.3021/241.95, showing that extreme A4
quantization benefits from reducing weight precision before activation
precision.  The intermediate teacher matters:
using the W1.58A32 teacher is stronger than using the full-precision teacher
under the same uniform ($\gamma=1$) budget-allocation schedule, improving FID from 3.64 to 3.39 and IS
from 227.23 to 244.00.  This supports the view that the replacement path should
follow the target quantization problem rather than simply
distill the final model from the full-precision network.

\textbf{Training path under stronger operators.}
Table~\ref{tab:robuq_training_path} shows that staged replacement improves
RobuQ, depending on the metric.  For
W4A4, Theseus QAD improves sFID from 5.20 to 4.48 and IS from 265.02 to
267.22, with a modest FID trade-off from 2.71 to 2.89.  For W1.58A4, it
improves all three metrics, reducing FID from 3.21 to 3.16 and sFID from 4.86
to 4.53 while increasing IS from 217.04 to 276.25.  This indicates that the
training strategy is compatible with operator-enhanced baselines, even though
its benefit is not uniformly expressed by every metric.

\vspace{-4mm}
\section{Conclusion}
\vspace{-3mm}
\label{sec:conclusion}

We presented \ours{}, a progressive distillation framework for deployable
4-bit-activation generative models, and \method{} as its core training method.
The central idea is to move the complexity required by aggressive A4
quantization from the inference graph into the training path.  By aligning
matched groups before returning to the original task objective, \method{} trains
plain W4A4 and W1.58A4 operators that would be unstable or less diverse under
direct replacement.  The DiT, FLUX, and DLLM results show that this approach
recovers low-bit quality in both controlled and data-limited settings while
keeping the deployed operator simple.

\textbf{Takeaway.} For smaller models, \method{} supplies the warm-up stage
missing from standard QAD, enabling subsequent end-to-end QAT of plain low-bit
models. For larger models, where small-batch QAT is unstable, \method{} is a
more effective distillation strategy than  QAD.

\textbf{Limitations and future work.} We have not reported experiments on
autoregressive LLMs, nor have we evaluated large models with sufficient data
and compute for extended end-to-end training. Testing whether the same
training path remains effective in these settings is an important next step.

\subsection*{AI Use Statement}
We used generative AI tools to assist with code development and language polishing.
The core tables, conclusions, and figures were prepared by the authors without
generative AI assistance. All AI-assisted content and code were reviewed and
verified by the authors. The authors take full responsibility for the final
content of this work, including any text or artifacts produced with the aid of
generative AI.

\subsection*{Ethics Statement}
This work studies quantization methods for improving the efficiency of diffusion
models. Quantized models may retain the biases and potential for misuse of their
pretrained counterparts, and lower deployment costs may make these models more
widely accessible. Our evaluations focus on generation quality and computational
efficiency and do not establish the safety of the quantized models.

\subsection*{Reproducibility Statement}
We will publicly release our code and model weights to facilitate reproducibility.
Detailed experimental settings and implementation details will be provided in
the appendix.

\bibliographystyle{arxiv_conference}
\bibliography{references}
\clearpage
\appendix
\section{Visual Comparisons}
\label{app:visual_comparisons}

We provide additional visual comparisons to complement the main-paper results.
Figure~\ref{fig:dit_w1p58a4_visual_comparisons} presents two panels of
DiT-XL/2 comparisons under W1.58A4 quantization, while
Figure~\ref{fig:flux_schnell_w4a4_visual_comparison} presents
FLUX.1-schnell comparisons under W4A4 INT4 quantization.

\begingroup
\setlength{\intextsep}{6pt}
\begin{figure}[H]
  \centering
  \includegraphics[width=0.95\textwidth]{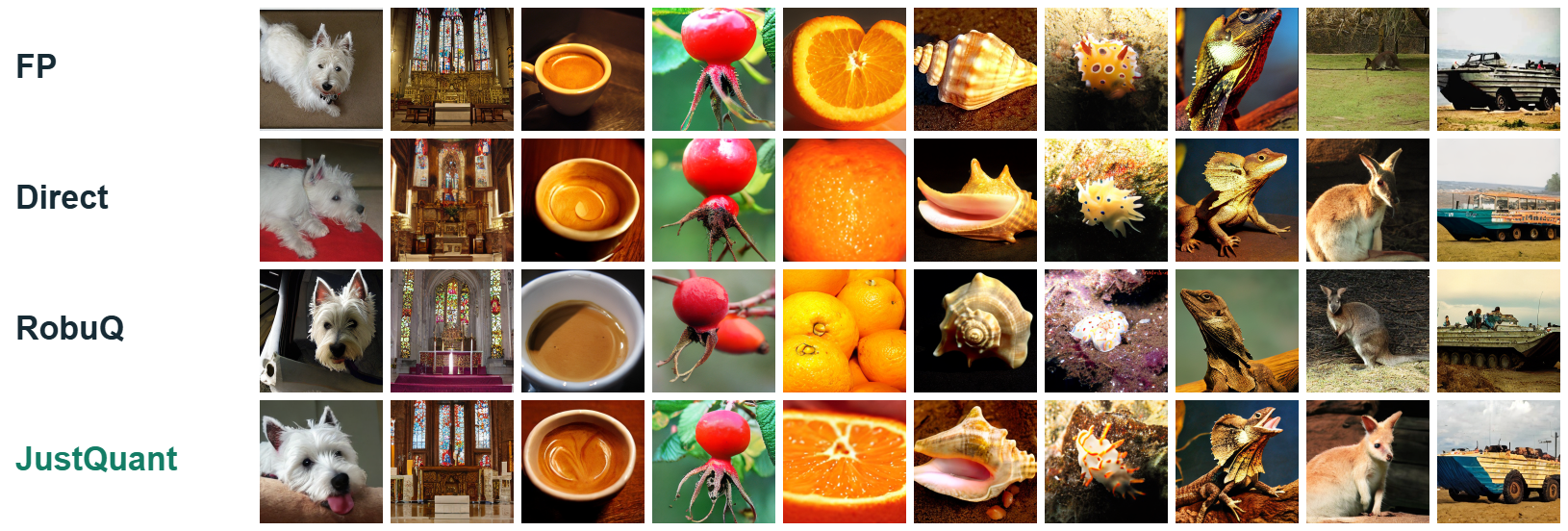}
  \vspace{1mm}
  \includegraphics[width=0.95\textwidth]{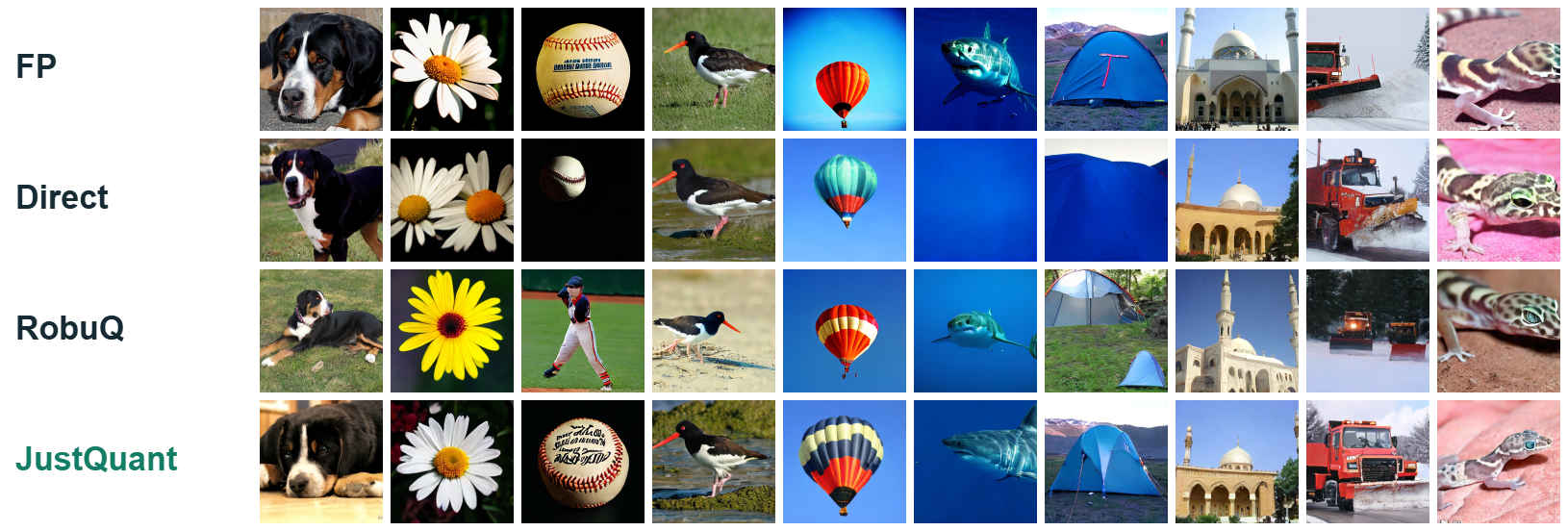}
  \caption{\textbf{DiT-XL/2 W1.58A4 qualitative comparisons.} The two panels
  show ImageNet $256\times256$ samples from the full-precision model and the
  compared low-bit methods under the W1.58A4 setting.}
  \label{fig:dit_w1p58a4_visual_comparisons}
\end{figure}

\begin{figure}[H]
  \centering
  \includegraphics[width=0.95\textwidth]{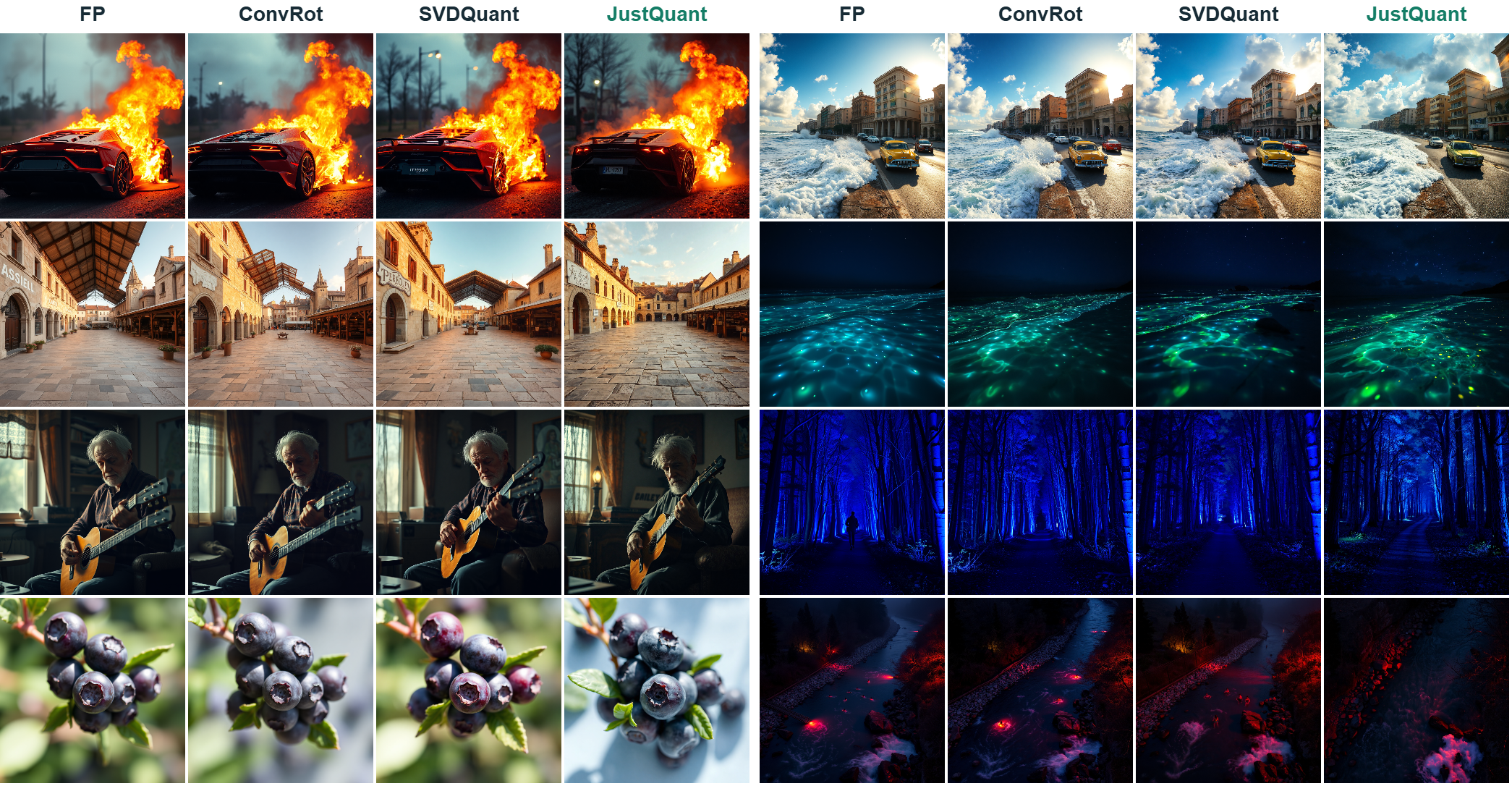}
  \caption{\textbf{FLUX.1-schnell W4A4 INT4 qualitative comparison.} Each row
  shows one prompt, and the columns compare FP, ConvRot, SVDQuant, and
  JustQuant under the W4A4 INT4 setting.}
  \label{fig:flux_schnell_w4a4_visual_comparison}
\end{figure}
\endgroup
\newpage

\section{Experimental Details}
\label{app:experimental_details}

This section records the configurations used for the experiments reported in
the main paper. Each column of the table is one experimental setting and each
row is one configuration field. Values marked as not recorded are not inferred
from the available experiment logs.

\begin{table*}[!htbp]
\centering
\caption{DiT-XL/2 experimental details.}
\label{tab:app_dit_experiment_details}
\scriptsize
\setlength{\tabcolsep}{3pt}
\renewcommand{\arraystretch}{1.18}
\begin{tabularx}{\textwidth}{@{}p{0.19\textwidth}*{3}{>{\centering\arraybackslash}X}@{}}
\toprule
 & \multicolumn{3}{c}{DiT-XL/2, ImageNet $256^2$} \\
\cmidrule(lr){2-4}
Setting & W1.58A32 & W4A4 & W1.58A4 \\
\midrule
\multicolumn{4}{l}{\textbf{Stage 1 (Theseus QAD)}} \\
Dataset & ImageNet, 128k VAE latents & ImageNet, 128k VAE latents & ImageNet, 128k VAE latents \\
Learning rate & $1\times10^{-4}$ & $1\times10^{-4}$ & $1\times10^{-4}$ \\
Batch size & 256 & 128 & 256 \\
Training steps & 10k & 10k & 10k \\

$\gamma$ (budget) & 1 & 2 & 2 \\
\midrule
\multicolumn{4}{l}{\textbf{Stage 2 (Official QAT)}} \\
Dataset & ImageNet, 128k VAE latents & ImageNet, 128k VAE latents & ImageNet, 128k VAE latents \\
Learning rate & $1\times10^{-4}$ & $1\times10^{-4}$ & $1\times10^{-4}$ \\
Batch size & 256 & 128 & 256 \\
Training steps & 100k & 10k & 100k \\
\midrule
GPU hours & 12 h $\times$ 8 H200 = 96 GPU-hours & 1.75 h $\times$ 4 H200 = 7 GPU-hours & 12 h $\times$ 8 H200 = 96 GPU-hours \\
Quantization function & Ternary weights, FP activations & INT4 weights/activations, group128 & Ternary weights, INT4 activations, group128 \\
\bottomrule
\end{tabularx}

\par\vspace{6pt}
\normalsize
\caption{FLUX experimental details.}
\label{tab:app_flux_experiment_details}
\scriptsize
\setlength{\tabcolsep}{3pt}
\renewcommand{\arraystretch}{1.18}
\begin{tabularx}{\textwidth}{@{}p{0.19\textwidth}*{2}{>{\centering\arraybackslash}X}@{}}
\toprule
 & \multicolumn{2}{c}{FLUX (W4A4)} \\
\cmidrule(lr){2-3}
Setting & FLUX-schnell & FLUX-dev \\
\midrule
\multicolumn{3}{l}{\textbf{Stage 1 (Theseus QAD)}} \\
Dataset & Pick-High 100K refined image-text subset~\citep{pick_high_dataset} & Pick-High 100K refined image-text subset~\citep{pick_high_dataset} \\
Learning rate & $1\times10^{-5}$ & $1\times10^{-5}$ \\
Batch size (global) & 16 & 16 \\
Training steps & 10k & 10k \\
Training epochs & 2 & 2 \\

$\gamma$ (budget) & 2 & 2 \\
\midrule
GPU hours & 18 h $\times$ 4 H200 = 72 GPU-hours & 18 h $\times$ 4 H200 = 72 GPU-hours \\
Quantization function & W4A4 INT4 block/group64 linears & W4A4 INT4 block/group64 linears \\
\bottomrule
\end{tabularx}

\par\vspace{6pt}
\normalsize
\caption{ELF-B and LLaDA-8B experimental details.}
\label{tab:app_elf_llada_experiment_details}
\scriptsize
\setlength{\tabcolsep}{3pt}
\renewcommand{\arraystretch}{1.18}
\begin{tabularx}{\textwidth}{@{}p{0.19\textwidth}*{3}{>{\centering\arraybackslash}X}@{}}
\toprule
 & \multicolumn{2}{c}{ELF-B} & LLaDA-8B \\
\cmidrule(lr){2-3}
Setting & W1.58A4 & W4A4 & W4A4 \\
\midrule
\multicolumn{4}{l}{\textbf{Stage 1 (Theseus QAD)}} \\
Dataset & OpenWebText & OpenWebText & GSM8K train \\
Learning rate & $2\times10^{-3}$, cosine $\rightarrow 1.25\times10^{-5}$ & $2\times10^{-3}$, cosine $\rightarrow 1.25\times10^{-5}$ & $1\times10^{-5}$ \\
Batch size & 512 effective & 512 effective & 8 \\
Training steps & 18,764 & 18,764 & 9.3k \\

$\gamma$ (budget) & 2 & 2 & 2 \\
\midrule
\multicolumn{4}{l}{\textbf{Stage 2 (Official QAT)}} \\
Dataset & OpenWebText & OpenWebText & N/A \\
Learning rate & $1\times10^{-5}$, constant & $1\times10^{-5}$, constant & N/A \\
Batch size & 512 effective & 512 effective & N/A \\
Training steps & 45,036 & 16,884 & N/A \\

\midrule
GPU hours & 19 h $\times$ 8 H200 = 152 GPU-hours  & 12 h $\times$ 8 H200 = 96 GPU-hours  & 2 h $\times$ 4 H200 = 8 GPU-hours \\

Quantization function & INT4 weights/activations, group128 & Ternary weights, INT4 activations, group128 & NVFP4 W4A4 \\
\bottomrule
\end{tabularx}
\end{table*}

 \section{Frequently Asked Questions and Clarifications}
  \label{app:faq}

  \paragraph{Q1: Why schedule distillation widths, and what does $\gamma$ control?}
  A single fixed width suited to one model need not transfer to
  other sizes or architectures. We use a binary-merge schedule,
  such as $(1,2,4,8,16,28)$ for DiT-XL/2, and allocate training
  steps across its levels using $N_k \propto \gamma^k$.
  Thus, $\gamma$ controls the budget, not the widths.
  $\gamma=1$ gives a uniform allocation; $\gamma=0$ corresponds
  to block-wise KD, while $\gamma=\infty$ corresponds to
  full-model distillation.

  \paragraph{Q2: What is progressive, and must the smallest unit be a block?}
  Student segments progressively grow by merging neighboring units,
  with parameters inherited across levels. Student blocks compose
  within each segment, but segment inputs remain teacher activations;
  student outputs are not passed between segments. The smallest
  unit need not be a full block: in FLUX, using MLP or attention
  modules yielded comparable performance.

  \paragraph{Q3: What evidence supports the short-to-long order?}
  Figure~\ref{fig:gamma_sweep} includes reverse-order results.
  For W4A4 with $\gamma=2$, reversing the order increases FID
  from 6.28 to 8.70; for W1.58A32 with $\gamma=1$, it increases
  FID from 12.87 to 29.50. These results support the short-to-long
  order in the tested settings, not its superiority over every
  possible schedule.

  \paragraph{Q4: Why not jointly optimize losses from all levels?}
  We observed that distillation losses at different block locations
  can differ in magnitude by up to five orders of magnitude.
  Joint multi-level supervision therefore requires careful,
  model-dependent loss weighting. We schedule supervision across
  levels to reduce this tuning burden and improve transferability.

  \paragraph{Q5: How does this differ from prior block-wise training?}
  Block-wise alignment is not our novelty. DilateQuant combines
  BKD with Weight Dilation and a Temporal Parallel Quantizer
  \citep{liu2024dilatequant}; EfficientQAT uses block-wise training
  followed by end-to-end quantization-step-size training
  \citep{chen2024efficientqat}. Our focus is progressively merging
  teacher-fed student segments and allocating steps across their
  lengths. The $\gamma=0$ setting provides the block-wise KD
  baseline within our training framework.

  \paragraph{Q6: Why use different training stages across models?}
  For smaller models, Theseus QAD provides an initialization for
  subsequent task-level QAT. For FLUX and LLaDA-8B, we use
  Theseus QAD alone under the available budget and small-batch
  constraints. This choice does not imply that larger models
  cannot benefit from additional QAT.

  \paragraph{Q7: Do plain operators eliminate the need for specialized kernels?}
  No. Low-bit execution still requires suitable quantization,
  packing, and matrix-multiplication kernels. ``Plain'' means
  avoiding additional smoothing, rotation, normalization, or SVD
  branches introduced for quantization, not eliminating
  low-bit kernel engineering.

  \paragraph{Q8: Does FLUX W4A4 cover the entire generation pipeline?}
  No. W4A4 applies to transformer-block linear layers.
  Text encoders, the VAE, and boundary modules remain in high
  precision. This transformer-focused scope follows
  SVDQuant~\citep{li2024svdquant}; it does not mean that all
  parameters or operations are four-bit.

  \paragraph{Q9: Why are multiple seeds not reported for every experiment?}
  Repeating every training and evaluation run across seeds would
  exceed our available compute budget. We prioritize multi-seed
  evaluation for ELF generation, where results are seed-sensitive.
  Results without repeated runs do not quantify run-to-run
  variability.
\end{document}